\documentclass{elsarticle}

\usepackage{xcolor}

\usepackage{booktabs}

\usepackage{mathtools}

\usepackage{amssymb}

\usepackage{bm}

\usepackage[T1]{fontenc}
\usepackage{lmodern}

\usepackage{tikz}
\usepackage{pgfplots}
\pgfplotsset{compat=1.15}
\pgfplotsset{every tick label/.append style={font=\small}}

\usepackage{hyperref}

\usepackage{caption}
\usepackage{subcaption}

\usepackage[ruled]{algorithm}
\usepackage{algorithmic}

\usepackage[%
    activate={true,nocompatibility},%
    final,tracking=true,%
    kerning=true,%
    spacing=true,%
    factor=1100,%
    stretch=10,%
shrink=10]{microtype}

\microtypecontext{spacing=nonfrench}

\DeclareMathOperator*{\argmin}{arg\,min} 
\makeatletter 
\let\oldtheequation\theequation
\renewcommand\tagform@[1]{\maketag@@@{\ignorespaces#1\unskip\@@italiccorr}}
\renewcommand\theequation{(\oldtheequation)} 
\makeatother

\newcommand{\ah}[1]{\textcolor{black}{#1}}
\newcommand{\rv}[1]{\textcolor{black}{#1}}

\begin{document}
\begin{frontmatter}
    \title{Physics informed neural networks for continuum micromechanics}
    \author{Alexander Henkes\corref{cor1}
    \fnref{label1, label2}}
    \ead{a.henkes@tu-braunschweig.de}
    \cortext[cor1]{Corresponding author}
    \fntext[label2]{https://orcid.org/0000-0003-4615-9271}
    \author{Henning Wessels%
    \fnref{label1}}
    \author{Rolf Mahnken%
    \fnref{label3}}
    \affiliation[label1]{
        organization={Institute for Computational Modeling in Civil Engineering,
        Technical University of Braunschweig},
        addressline={Pockelsstr. 3}, 
        city={Braunschweig},
        postcode={38106}, 
    country={Germany}}

    \affiliation[label3]{
        organization={Chair of Engineering Mechanics, University of Paderborn},
        addressline={Warburger Str. 100}, 
        city={Paderborn},
        postcode={33098}, 
    country={Germany}}

    \begin{abstract}
        Recently, physics informed neural networks have successfully been
        applied to a broad variety of problems in applied mathematics and
        engineering. The principle idea is \ah{the usage of} a neural network as
        a global ansatz function \ah{for} partial differential equations. Due to
        the global approximation, physics informed neural networks have
        difficulties in displaying localized effects and strong \ah{nonlinear
        solution fields} by optimization. In this work we consider \ah{nonlinear
            stress and displacement fields} invoked by material inhomogeneities
            with sharp phase interfaces. This constitutes a challenging problem
            for a method relying on a global ansatz. To overcome convergence
            issues, adaptive training strategies and domain decomposition are
            studied. It is shown, that the domain decomposition approach is
            \ah{capable} to accurately resolve nonlinear stress, displacement
            and energy fields in heterogeneous microstructures obtained from
            real-world $\mu$CT-scans.
    \end{abstract}
%
%
%

    \setcounter{tocdepth}{3}

\end{frontmatter}
\section{Introduction} \noindent
Problems in continuum mechanics arise in many areas of engineering, like civil-,
mechanical- and aerospace engineering. The development of hybrid materials such
as fiber reinforced plastics offers great potential for lightweight design due
to their high strength to density ratio, compared to classic engineering
materials. Unfortunately, these materials also raise the computational effort
of meshing and simulation, as their overall macroscopic behavior is defined by
their underlying heterogeneous microstructure. In order to resolve tensor fields
of interest, like displacement, stress and strain fields, numerical methods have
to be used in absence of analytical solutions. The most widely used numerical
method, especially in computational mechanics, is the \textit{finite element
method} (FEM). Despite \ah{its} broad range of applications and the successes
therein, open problems remain. Among these, multi-scale problems, uncertainty
quantification and inverse problems impose challenges to FEM. The reader is
referred to \cite{david2020application} for a recent review.

An alternative to conventional numerical methods are \textit{artificial
neural networks} (ANN). ANNs are able to solve a wide range of problems,
such as computer vision, speech recognition and autonomous driving
\cite{lecun2015deep}.
In the scientific context, ANNs gain more and more popularity. They were
successfully applied in fields like quantum mechanics \cite{lantz2021deep},
bioinformatics \cite{min2017deep}, medicine \cite{piccialli2021survey} and
applied mathematics \cite{lu2021learning}.
Recently, several investigations were made in the field of continuum
mechanics \cite{bock2019review}. Applications reach from fluid mechanics
\cite{kutz2017deep} over fracture mechanics \cite{hsu2020using} to
micromechanics in the context of uncertainty quantification of effective
properties \cite{henkes2021deep}.  

While traditional ANN approaches need large datasets for training, the discovery
of \textit{physics informed neural networks} (PINNs) provides a framework for
small data regimes. PINNs are introduced in \cite{lagaris1998artificial} and
have been \ah{reinvented} using automatic differentiation and application to
time-dependent problems in \cite{raissi2019physics}. Here, the amount of data
needed to solve scientific problems can even be reduced to initial and boundary
conditions by solving the underlying governing equations, \ah{which are} often
described by
\textit{partial differential equations} (PDEs) \cite{karniadakis2021physics}.
PINNs therefore present an alternative to established numerical methods like
FEM.  They were successfully applied to bioinformatics
\cite{rackauckas2020universal}, power systems \cite{misyris2020physics} and
chemistry \cite{ji2020stiff}, among other fields.

Like in the case of standard ANNs, several contributions of PINNs to continuum
mechanics were published. 
Among these, fluid mechanics were considered in the context of flow
visualization \cite{raissi2020hidden}, multi-physics additive manufacturing
simulations \cite{zhu2021machine} and free surface flows
\cite{wessels2020neural}, as well as for numerous other applications, see
\cite{cai2021physicsfluid} for a review.
In solid mechanics, PINNs were applied to solve problems in curing
\cite{niaki2021physics}, material identification \cite{zhang2020physics}, heat
transfer \cite{cai2021physics}, forward problems in plate theory
\cite{vahab2021physics}, plasticity \cite{haghighat2021physics}, elastodynamics
\cite{rao2020physics} and surrogate
modeling \cite{haghighat2021physics}.  Elastostatics were tackled using energy
based methods for linear elasticity \cite{samaniego2020energy} and
hyperelasticity \cite{kollmannsberger2021deep}.  Energy based error bounds can
be found in \cite{guo2020energy}. The reconstruction of material distributions
from given strain fields was the topic of  \cite{chen2021learning}.

While the effectiveness of the PINN approach in applications to continuum
mechanics in the context of homogeneous materials was shown in the literature,
to the best of the authors knowledge no attempts were made to resolve
displacement and stress fields resulting from inhomogeneous material
distributions using PINNs directly without the need \ah{of} additional data. In
the context of micromechanics, the knowledge of the exact micro stress field is
crucial for full field homogenization. Unfortunately, several problems arise in
this context, such as jumps in material parameters, highly nonlinear solution
fields and convergence. The convergence to an optimal solution in the context of
PINN was proved for linear PDEs \cite{shin2020convergence} but \ah{the
performance of these techniques} \rv{in inhomogeneous
cases remains open. In this context, classical techniques known from standard
numerical methods, like \textit{h}- and \textit{p}-refinement
(capacity of the ANN), are investigated for PINNs.} To gain
confidence in the ability of PINNs to capture the underlying physics of
inhomogeneous micromechanics, the present study therefore aims towards the
following key contributions: 
\begin{itemize} 
    \item \textbf{PINN elastostatics of inhomogeneous materials:}
        A PINN for calculating displacement, stress and energy fields for
        arbitrary microstructural unit cells with arbitrary material parameters
        is proposed. The topology choices, loss calculation, boundary
        conditions and scaling methods are discussed in detail.     
    \item \textbf{Adaptivity and domain decomposition:} 
        To improve convergence, an adaptive collocation point sampling
        algorithm \ah{(\textit{h}-refinement)} as well as a domain decomposition
        technique \ah{(\textit{p}-refinement)} are discussed. 
    \item \textbf{$\bm{\mu}$CT-scan of wood-plastic composite:}
        The proposed method's performance is investigated on a real-world
        $\mu$CT-scan of a wood-plastic composite. To resolve the
        microstructure, a material network is introduced to render a smooth
        extension of voxelized image data. It is shown, that the PINN approach
        is able to accurately solve the micromechanical boundary value problem
        (BVP) of inhomogeneous elastostatics.
\end{itemize}
The proposed PINN solves the strong form of the underlying BVP directly,
\ah{thus it allows} to work with voxelized image data in a straightforward
manner.  \ah{Moreover,} the formulation is flexible, making it easy to add
additional physical or experimental information into the solution process. This
opens up \ah{the application of this method} in uncertainty quantification and
inverse problems, such as the identification of uncertain material parameters.
\rv{This is in contrast to the Deep Energy Method (DEM)
    \cite{samaniego2020energy, nguyen2020deep}, where the principal of
    stationary elastic potential or \textit{energy approach} was used for
    homogeneous problems in computational mechanics. DEM relies on integration
    of the energy formulation and uses a regular grid. Also, it is not
    well-suited for inverse heterogeneous problems, as the energy, used as
    optimization variable, is a global measure. Therefore, two local stress
    fields can lead to the same non-unique global energy value.  Furthermore,
the trivial solution always leads to a minimum of the energy, thus prohibiting
the identification of parameters.}

The remainder of this paper is structured as follows. In
\autoref{sec:governing_equations}, the governing equations of linear elastic
micromechanics are reviewed. 
\autoref{sec:discretization} summarizes the basic equations and concepts of ANN
and PINN. In \autoref{sec:pinn_meca}, aspects of PINN discretization in the
context of linear elastic micromechanics, including discussions about network
topology, loss function calculation, boundary condition handling and scaling
methods. The section closes with two numerical examples. Then, advanced
techniques such as adaptivity and domain decomposition are developed in
\autoref{sec:advanched}. A convergence study, comparing all proposed methods, is
conducted. The best performing techniques are then used to calculate micro
displacement and micro stress fields of a real-world $\mu$CT-scan of a
wood-plastic composite in \autoref{sec:ct_scan}. To \ah{reach} a smooth
extension of the \ah{discrete voxelized} image resulting from $\mu$CT-scans, a
material network is proposed.  The paper closes with a summary and an outlook in
\autoref{sec:conc_out}.

\section{Governing equations of linear elastic micromechanics}
\label{sec:governing_equations}
\noindent
In this section, a short overview of the governing equations of micromechanics,
\ah{following \cite{li2008introduction}
is given.} The PDEs considered in this work
arise in the context of elastostatics, assuming linear elastic material
behaviour. To this end, the \textit{domain} of interest $\Omega$ is chosen to be
a symmetric and zero-centered \rv{square} \textit{unit cell}, such that
\begin{equation}
    \Omega = \left\{\bm{x} = \left\{ x^d \right\} \in \mathbb{R}^d \;\bigg| \;
    \frac{-L}{2} \leq x^d \leq \frac{L}{2} \right\}.
    \label{eq:domain}
\end{equation}
Here, $\bm{x}$ are points in the domain $\Omega$ and $L$ denotes the
\textit{edge length} of the unit cell $\Omega$.  The boundary of the domain
$\Omega$ is denoted by 
\begin{equation}
    \partial \Omega = \partial \Omega_{\bar{u}} \cup \partial \Omega_{\bar{t}}, 
    \quad 
    \partial \Omega_{\bar{u}} \cap \partial \Omega_{\bar{t}} = \emptyset,
\end{equation}
where $\Omega_{\bar{u}}$ are sections subject to \textit{Dirichlet boundary
conditions}
\begin{equation}
    \bm{u}(\bm{x}) = 
    \bar{\bm{u}}(\bm{x}),\quad \bm{x} 
    \in \partial \Omega,
    \label{eq:disp}
\end{equation}
where $\bm{u}(\bm{x})$ is a \textit{displacement vector field} and
$\bar{\bm{u}}(\bm{x})$ represent \textit{prescribed displacements}. Furthermore,
$\partial\Omega_{\bar{t}}$ is subject to \textit{Neumann boundary conditions} 
\begin{equation}
    \bm{t}(\bm{x}) = {\bm{\sigma}}(\bm{x})
    \cdot \bm{n}(\bm{x})= \bar{\bm{t}}(\bm{x})
    = \bar{\bm{\sigma}}(\bm{x}) \cdot \bm{n}(\bm{x}),
    \quad \bm{x} \in \partial \Omega,
    \label{eq:trac}
\end{equation}
with \textit{traction} $\bm{t}(\bm{x})$, \textit{stress tensor field}
${\bm{\sigma}}(\bm{x})$, \textit{normal vector} $\bm{n}(\bm{x})$ and
\textit{prescribed traction} $\bar{\bm{t}}(\bm{x})$. The governing equations
include the balance law of linear momentum
\begin{equation}
    \nabla \cdot \bm{\sigma}(\bm{x}) = 0,
    \quad \bm{x} \in \Omega,
    \label{eq:lin_mom}
\end{equation}
excluding body forces and dynamic terms. Symmetry of the stress tensor 
$\bm{\sigma}(\bm{x})$ guarantees the balance of angular momentum
\begin{equation} 
    \bm{\sigma}(\bm{x})= 
    \bm{\sigma}^T(\bm{x}),\quad \bm{x} \in \Omega.
    \label{eq:angular}
\end{equation}
The displacement vector field $\bm{u}(\bm{x})$ can be related to a strain tensor 
field $\bm\varepsilon(\bm{x})$ by the kinematic relation
\begin{equation}
    \bm\varepsilon(\bm{x}) 
    = \frac{1}{2} \left(\nabla \bm{u}
    (\bm{x}) + \nabla \bm{u}^T(\bm{x}) \right),\quad \bm{x} \in \Omega.
    \label{eq:kin}
\end{equation}
The stress tensor field ${\bm{\sigma}}(\bm{x})$ and the strain tensor field 
$\bm\varepsilon(\bm{x})$ are related by a \textit{constitutive relation} or
\textit{material law}. In this work, linear elasticity is investigated, \ah{in
which}
the material law takes the form
\begin{equation}
    \bm{\sigma }(\bm{x}) = \lambda(\bm{x}) 
    \operatorname{tr}({\bm{\varepsilon}}(\bm{x}))
    \mathbb{I} + 2 \mu(\bm{x}) \bm{\varepsilon}(\bm{x}),\quad \bm{x} \in \Omega,
    \label{eq:const}
\end{equation}
where $\lambda(\bm{x})$ and $\mu(\bm{x})$ are scalar fields,
representing \textit{Lam\'{e}'s first and second constant}, respectively
\ah{and}
$\mathbb{I}$ denotes the \textit{second order identity tensor}. Finally, the
balance of \textit{internal work}
\begin{equation}
    W_{int} = \displaystyle\frac{1}{2} \displaystyle\int_{\Omega} 
    \bm{\varepsilon}(\bm{x}) : 
    \bm{\sigma}(\bm{x}) \mathrm{d \Omega},
\end{equation}
and \textit{external work}
\begin{equation}
    W_{ext} = 
    \displaystyle\int_{\partial \Omega} \bar{\bm{t}}(\bm{x})
    \cdot \bm{u}(\bm{x})  
    \mathrm{d \partial \Omega} =
    \displaystyle\int_{\partial \Omega} \bar{\bm{\sigma}}
    (\bm{x}) \cdot \bm{n}(\bm{x}) \cdot
    \bm{u}(\bm{x}) \mathrm{d \partial \Omega},
\end{equation}
can be expressed by $W_{int} = W_{ext}$ and
\begin{equation}
    \displaystyle\frac{1}{2} \displaystyle\int_{\Omega} 
    \bm{\varepsilon}(\bm{x}) : 
    \bm{\sigma}(\bm{x}) \mathrm{d \Omega} = 
    \displaystyle\int_{\partial \Omega} \bar{\bm{t}}(\bm{x})
    \cdot \bm{u}(\bm{x})  
    \mathrm{d \partial \Omega} =
    \displaystyle\int_{\partial \Omega} \bar{\bm{\sigma}}
    (\bm{x}) \cdot \bm{n}(\bm{x}) \cdot
    \bm{u}(\bm{x}) \mathrm{d \partial \Omega}. 
    \label{eq:energy}
\end{equation}
The primary variable of the micromechanical balance of linear momentum in
\autoref{eq:lin_mom} is the displacement vector $\bm{u}(\bm{x})$ introduced in
\autoref{eq:disp}. By inserting the material law in \autoref{eq:const} as well
as the kinematic relation from \autoref{eq:kin} into the balance law
\autoref{eq:lin_mom}, one
obtains the \textit{Navier-Cauchy equations} \ah{for homogeneous materials}
\begin{equation}
    \mu\Delta\bm{u}(\bm{x}) 
    + (\lambda+\mu)\nabla(\nabla\cdot\bm{u}(\bm{x})) = 0.
    \label{eq:navier_cauchy}
\end{equation}
For inhomogeneous materials, \ah{inserting the material law
\autoref{eq:navier_cauchy} and the kinematic
relation \autoref{eq:kin} into the balance law \autoref{eq:lin_mom} leads to}
\begin{equation}
    \nabla \cdot \left( \lambda(\bm{x}) 
        \operatorname{tr}\left( \frac{1}{2} \left(\nabla \bm{u}
        (\bm{x}) + \nabla \bm{u}^T(\bm{x})\right)\right)
        \mathbb{I} + \mu(\bm{x}) (\nabla \bm{u}
    (\bm{x}) + \nabla \bm{u}^T(\bm{x})) \right) = 0,
    \label{eq:navier_cauchy_inhomo}
\end{equation}
due to chained derivatives.

\section{Artificial neural networks and physics informed neural networks}
\label{sec:discretization} \noindent
Whereas traditional numerical methods like FEM
are often concerned with local \ah{ansatz functions, PINNs use global ansatz
functions to find a global solution}
for the given BVP. This results in a complex optimization process, solved during
\textit{training} of an ANN. In the following, ANNs are concisely introduced in
\autoref{sec:ann} and the theory of PINNs is explained in \autoref{sec:pinn}. In
\autoref{sec:pinn_meca}, PINNs are then used to discretize and solve BVPs
appearing in \autoref{sec:governing_equations}.
\subsection{Artificial neural networks} 
\label{sec:ann} \noindent
\rv{This section introduces the notation used in the context of ANNs. The reader
familiar with this topic may proceed directly with \autoref{sec:pinn}.}

\begin{figure}[htb] 
    \centering 
    \tikzset{%
        every neuron/.style={ circle,
        draw, minimum size=0.5cm }, neuron missing/.style={ draw=none, scale=1,
    execute at begin node=\color{black}$\vdots$ }, } 
    \begin{tikzpicture}[x=1cm,y=1cm, >=stealth] 
        \node at (0.1,-0.75) [] {$\bm{h}^{(0)}$}; \node at (2.1,-0.75) []
        {$\bm{h}^{(l)}$}; \node at (4.3,-0.75) [] {$\bm{h}^{(n_L)}$}; 

        \foreach \m/\l [count=\y] in {1, missing, 2} \node [every
        neuron/.try, neuron \m/.try] (input-\m) at (0,-1-\y) {};

        \foreach \m [count=\y] in {1, missing, 2, missing,3} \node [every
        neuron/.try, neuron \m/.try ] (hidden-\m) at (1,-0.9-\y*0.7) {};

        \foreach \m [count=\y] in {1, missing, 2, missing,3} \node [every
        neuron/.try, neuron \m/.try ] (hidden2-\m) at (2,-0.9-\y*0.7) {};

        \foreach \m [count=\y] in {1, missing, 2, missing,3} \node [every
        neuron/.try, neuron \m/.try ] (hidden3-\m) at (3,-0.9-\y*0.7) {};

        \foreach \m [count=\y] in {1, missing, 2} \node [every neuron/.try,
        neuron \m/.try ] (output-\m) at (4,-1-\y) {};

        \draw [<-] (input-1) -- ++(-1,0) node [above, midway] {$x_1$}; \draw
        [<-] (input-2) -- ++(-1,0) node [above, midway] {$x_{d_x}$};

        \draw [->] (output-1) -- ++(1,0) node [above, midway] {$y_1$}; \draw
        [->] (output-2) -- ++(1,0) node [above, midway] {$y_{d_y}$};

        \foreach \i in {1,...,2} \foreach \j in {1,...,3} \draw [->]
        (input-\i) -- (hidden-\j);

        \foreach \i in {1,...,3} \foreach \j in {1,...,3} \draw [->]
        (hidden-\i) -- (hidden2-\j);

        \node at (2.5,-1.6) [] {$\cdots$}; \node at (2.5,-3.0) []
        {$\cdots$}; \node at (2.5,-4.4) [] {$\cdots$};

        \foreach \i in {1,...,3} \foreach \j in {1,...,2} \draw [->]
    (hidden3-\i) -- (output-\j); \end{tikzpicture} 
    \caption{Feed forward neural network topology of an ANN $\mathcal{N}(\bm{x};
    \bm{\theta})$ as described in \autoref{eq:ann}.} 
    \label{fig:ann} 
\end{figure}
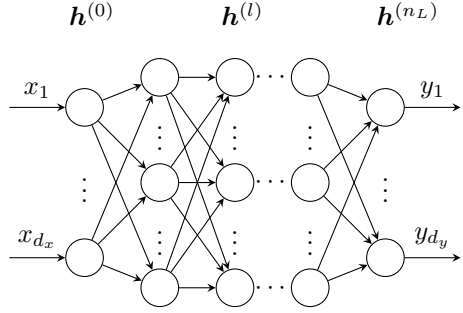

An ANN is a parametrized, nonlinear function composition. By the
\textit{universal function approximation theorem} \cite{hornik1989multilayer},
arbitrary Borel measurable functions can be approximated by an ANN. There are
several different formulations for ANN, which can be found in standard
references such as \cite{bishop2006pattern, goodfellow2016deep,
aggarwal2018neural, geron2019hands, chollet2018deep}.  In this paper,
\ah{\textit{densely
connected feed forward neural networks}} are used, which will be denoted by
$\mathcal{N}$. For the \ah{reminder of this contribution} the abbreviation ANN
is understood as
referring to such \ah{type of neural network.}
An ANN $\mathcal{N}$ is a function from an \textit{input space}
$\mathbb{R}^{d_x}$ to an \textit{output space} $\mathbb{R}^{d_y}$, defined by a
composition of nonlinear functions, such that \begin{align} \mathcal{N}:
    \mathbb{R}^{d_x} &\to \mathbb{R}^{d_y} \nonumber \\ \bm{x} &\mapsto
    \mathcal{N}(\bm{x}) = \bm{h}^{(l)} \circ \ldots \circ \bm{h}^{(0)} = \bm{y},
\quad l = 1, \ldots, n_L.  \label{eq:ann} \end{align} 
Here, $\bm{x}$ denotes an \textit{input vector} of dimension $d_x$ and $\bm{y}$
an \textit{output vector} of dimension $d_y$. The nonlinear functions
$\bm{h}^{(l)}$ are called \textit{layers} and define a $l$-fold composition,
mapping input vectors \ah{$\bm{x}$} to output vectors \ah{$\bm{y}$}.
Consequently, the first layer $\bm{h}^{(0)}$ is defined as the \textit{input
layer} and the last layer \ah{$\bm{h}^{(n_L)}$} as the \textit{output layer},
such
that \begin{equation} \bm{h}^{(0)} = \bm{x} \in \mathbb{R}^{d_x}, \qquad
\bm{h}^{(n_L)} = \bm{y} \in \mathbb{R}^{d_y}.  \label{eq:layer} \end{equation}
The layers \ah{$\bm{h}^{(l)}$} between the input and output layer, called
\textit{hidden layers},
are defined as
\begin{equation} \bm{h}^{(l)} = \left\{h_{\eta}^{(l)}, \; \eta = 1, \ldots,
    n_{u}\right\}, \qquad h_{\eta}^{(l)} =
    \phi^{(l)}\left(\bm{W}^{(l)}_{\eta}\bm{h}^{(l-1)}\right) =
\phi^{(l)}\left({z}_{\eta}^{(l)}\right), \label{eq:hidden} \end{equation} where
$h_{\eta}^{(l)}$ is the $\eta$-th \textit{neural unit} of the $l$-th layer
$\bm{h}^{(l)}$ and $n_u$ is the \textit{total number of neural units per layer},
which in this work is constant over all layers. $\phi^{(l)}$ is a nonlinear
\textit{activation function} from $\mathbb{R} \to \mathbb{R}$, which in this
work is defined by means of the so called
\textit{Swish} function \cite{ramachandran2017searching} as \begin{equation}
    \phi^{(l)}\left({z}_{\eta}^{(l)}\right) =
    \operatorname{swish}\left({z}_{\eta}^{(l)}\right) =
    \displaystyle\frac{1}{1+e^{-\beta {z}_{\eta}^{(l)}}}, \quad
    {z}_{\eta}^{(l)},
    \beta \in \mathbb{R} 
    \label{eq:activation}
\end{equation} with \begin{equation}
    \phi^{(n_L)}\left({z}_{\eta}^{(n_L)}\right) =
\operatorname{id}\left({z}_{\eta}^{(n_L)}\right).  \end{equation} Here,
${z}_{\eta}^{(l)}$ denotes the scalar valued product of the \textit{weight
vector} $\bm{W}_{\eta}^{(l)}$ of the $\eta$-th neural unit in the $l$-th layer
$\bm{h}^{(l)}$ and the output of the preceding layer $\bm{h}^{(l-1)}$ in
\autoref{eq:hidden}. In our notation we absorb the bias term, occurring in the
standard notation, as explained, \ah{ e.g.,} in \cite{aggarwal2018neural}. All
weight vectors $\bm{W}_{\eta}^{(l)}$ of all layers $\bm{h}^{(l)}$ can be
gathered in a single expression, such that 
\begin{equation}
    \bm{\theta}=\left\{\bm{W}_{\eta}^{(l)}\right\}, 
    \label{eq:parameters}
\end{equation} 
where $\bm{\theta}$ inherits all parameters of the ANN $\mathcal{N}(\bm{x})$
from \autoref{eq:ann}. Consequently, the notation $\mathcal{N}(\bm{x};
\bm{\theta})$ emphasises the dependency of the outcome of an ANN on the input on
the one hand and the current realization of the weights on the other hand. A
graphical representation of an ANN is shown in \autoref{fig:ann}, where the
specific combination of layers \ah{$\bm{h}^{(l)}$ and neural units
$h_{\eta}^{(l)}$ from \autoref{eq:hidden}}
and activation functions $\phi^{(l)}$
from \autoref{eq:activation} is called \textit{topology} of the ANN 
$\mathcal{N}(\bm{x}; \bm{\theta})$. 

\subsection{Physics informed neural networks} 
\label{sec:pinn} \noindent
An ANN $\mathcal{N}(\bm{x}; \bm{\theta})$, as described in \autoref{sec:ann},
can be used to solve PDEs by means of an optimization problem. This approach is
called PINNs \cite{raissi2019physics}.  In this work, stationary elliptic
nonlinear PDEs are considered, \ah{despite the fact that} PINNs are applicable
in more general
settings as well \cite{karniadakis2021physics}.  Following the notation from
\cite{berg2018unified}, consider a BVP over a domain $\Omega$ with boundary
$\partial \Omega$ defined by \begin{equation} \bm{Lu}(\bm{x}) = \bm{f}(\bm{x}),
    \bm{x} \in \Omega \subset \mathbb{R}^n, 
    \qquad \bm{Bu}(\bm{x}) = \bm{g}(\bm{x}),
\bm{x} \in \Gamma \subset \partial \Omega, \label{eq:bvp} \end{equation} where
$\bm{L}$ denotes a \textit{differential operator}, $\bm{u}(\bm{x})$ is a real
\textit{multivariate function} of a real \textit{multivariate variable} $\bm{x}$
of dimension $d$, representing spatial points of a domain $\Omega$, and
$\bm{f}(\bm{x})$ is a \textit{forcing term}.  Furthermore, $\bm{B}$ is a
\textit{boundary operator} with respective \textit{boundary data}
$\bm{g}(\bm{x})$, which is imposed on the corresponding part $\Gamma$ of the
boundary. In PINNs, the ansatz for solving \autoref{eq:bvp} is defined by means
of an ANN in \autoref{eq:ann} as \begin{equation} \bm{u}(\bm{x}) \approx
\mathcal{N}(\bm{x}; \bm{\theta}).  \label{eq:ansatz} \end{equation} Because an
ANN $\mathcal{N}(\bm{x}; \bm{\theta})$ can approximate arbitrary functions, it
is capable of representing the solution of the BVP given appropriate
parameters $\bm{\theta}$ from \autoref{eq:parameters}.

\subsubsection{Optimization} 
\label{sec:optimization} \noindent
The appropriate parameters $\bm{\theta}$ in \autoref{eq:parameters} of the ANN
$\mathcal{N}(\bm{x}; \bm{\theta})$ from \autoref{eq:ann}) can be found by
utilizing a collocation method. The domain $\Omega$ and the boundary part
$\Gamma$ from \autoref{eq:bvp} are discretized into sets of \textit{collocation
points} $\Omega_d$ and $\Gamma_d$, respectively,  with $|\Omega_d| = n_d$ and
$|\Gamma_d| = n_b$. Then, an optimization problem to find the optimal parameters
$\bm{\theta}^*$, also called \textit{training}, is defined as 
\begin{equation}
    \bm{\theta}^* = \underset{\bm{\theta}}\argmin\;
    \mathcal{L}(\bm{x}; \bm{\theta}) =
    \underset{\bm{\theta}}\argmin\; \left(\mathcal{L}_r(\bm{x}; \bm{\theta}) 
    + \mathcal{L}_b(\bm{x}; \bm{\theta})\right),
    \label{eq:optimization} 
\end{equation} 
with 
\begin{equation} \mathcal{L}_r(\bm{x}; \bm{\theta}) =
    \displaystyle\sum_{\bm{x} \in \Omega_d} \displaystyle\frac{1}{n_d} ||\bm{L}
    \mathcal{N}(\bm{x}; \bm{\theta}) - \bm{f}(\bm{x})||_2^2, \label{eq:loss_r}
\end{equation} 
and 
\begin{equation} 
    \mathcal{L}_b(\bm{x}; \bm{\theta}) = \displaystyle\sum_{\bm{x}
    \in \Gamma_d} \displaystyle\frac{1}{n_b} ||\bm{B} \mathcal{N}(\bm{x};
    \bm{\theta}) - \bm{g}(\bm{x})||_2^2, 
    \label{eq:loss_b} 
\end{equation} 
where $\bm{\theta}^*$ are the \textit{optimal weights and biases} with respect
to the objective function $\mathcal{L}(\bm{x}; \bm{\theta})$. The expression
$\mathcal{L}(\bm{x}; \bm{\theta})$ in \autoref{eq:optimization} is also called
\textit{loss function} and contains a residual term $\mathcal{L}_r(\bm{x};
\bm{\theta})$ of the PDE in \autoref{eq:loss_r} as well as a term
$\mathcal{L}_b(\bm{x}; \bm{\theta})$ in \autoref{eq:loss_b}, describing the
discrepancy of the boundary conditions. The spatial derivatives in the loss
function can be obtained by automatic differentiation of the ANN
$\mathcal{N}(\bm{x}; \bm{\theta})$ using open source frameworks like
\textit{Tensorflow} \cite{tensorflow2015-whitepaper}. The automatic
differentiation of \autoref{eq:ann} with respect to the spatial coordinates
implies the chain rule. 

\subsubsection{Boundary conditions} \noindent
The formulation of the optimization problem
in \autoref{eq:optimization} enforces the boundary conditions in
\autoref{eq:bvp} by means of a separate loss term $\mathcal{L}_b(\bm{x};
\bm{\theta})$. This approach
is called \textit{soft boundary conditions} in PINN terminology
\cite{sun2020surrogate}. The boundary conditions are fulfilled, if the loss term
$\mathcal{L}_b(\bm{x}; \bm{\theta})$ vanishes. On the other hand, so called
\textit{hard boundary
conditions} include the boundary conditions in the formulation of the
ANN in \autoref{eq:ann} a priori. Following the approach in
\cite{berg2018unified}, hard boundary conditions can be realized by letting the
boundary operator $\bm{B}$ from \autoref{eq:bvp} be the identity operator. Then,
the ansatz is formulated as 
\begin{equation} 
    {\mathcal{N}}(\bm{x}; \bm{\theta})
    = \bm{G}(\bm{x}) + \bm{D}(\bm{x}) \tilde{\mathcal{N}}(\bm{x}; \bm{\theta}),
    \label{eq:hard_bc} 
\end{equation} 
where $\bm{G}(\bm{x})$ is a \textit{smooth extension of the boundary data}
$\bm{g}(\bm{x})$ in \autoref{eq:bvp} and $\bm{D}(\bm{x})$ is a \textit{smooth
distance} function. $\tilde{\mathcal{N}}(\bm{x}; \bm{\theta})$ denotes an ANN
without boundary condition restrictions. If the distance function
$\bm{D}(\bm{x})$ vanishes, the point $\bm{x}$ lies on the boundary and the
output of the ANN is automatically \ah{neglected}. \ah{The functions}
$\bm{G}(\bm{x})$ and $\bm{D}(\bm{x})$ can be given analytically or represented
by ANNs. The advantage of hard boundary conditions over soft boundary conditions
is the reduced complexity of the optimization problem in
\autoref{eq:optimization}, which simplifies to
\begin{equation} \bm{\theta}^* =
    \underset{\bm{\theta}}\argmin\; \mathcal{L}(\bm{x}; \bm{\theta}) =
    \underset{\bm{\theta}}\argmin\; \mathcal{L}_r(\bm{x}; \bm{\theta}).  
    \label{eq:optimization2}
\end{equation} 
The loss function consists only of the residual loss
$\mathcal{L}_r(\bm{x}; \bm{\theta})$ in \autoref{eq:loss_r} with respect to the
PDE in
\autoref{eq:bvp}.

\section{Physics informed neural networks for linear elasticity}
\label{sec:pinn_meca}
\label{sec:topology} \noindent
\rv{
    In this work, the linear elastostatic balance law of linear momentum
presented in \autoref{eq:lin_mom} is directly solved using its strong form.}

To solve BVP in linear elasticity, a suitable topology for the ANN in
\autoref{eq:ann} and consequently the PINN described in \autoref{sec:pinn} has
to be chosen. One possibility is to let the output layer $\bm{h}^{(n_L)}$ of the
ANN be the tuple of the displacement vector field components $\bm{u}(\bm{x})$ in
\autoref{eq:disp}, i.e. the primary variables, such that
\begin{equation}
    \mathcal{N}(\bm{x}; \bm{\theta}) 
    = \left\{ \mathcal{N}_{u_{x}}(\bm{x}; \bm{\theta}), 
    \mathcal{N}_{u_{y}}(\bm{x}; \bm{\theta})\right\}.
    \label{eq:pinn_u}
\end{equation}
Then, the ANN can be used to solve the inhomogeneous Navier-Cauchy equation
in \autoref{eq:navier_cauchy_inhomo}, such as e.g., in \cite{nguyen2020deep}.
Unfortunately, several problems arise along utilization of such an approach.
First, in \autoref{eq:navier_cauchy_inhomo} the derivatives of the material
parameters $\lambda(\bm{x})$ and $\mu(\bm{x})$ are needed. For sharp material
phase transitions, numerical problems can arise during optimization in
\autoref{eq:optimization} due to exploding gradients. Second, in
\autoref{eq:navier_cauchy_inhomo} second derivatives of the displacement vector
$\bm{u}(\bm{x})$ are needed. The computation of these
introduces additional overhead during optimization.
Third, the application of mixed boundary conditions renders more difficult, as
e.g. free boundary stress conditions cannot be imposed in a straightforward
fashion using hard boundary conditions described in \autoref{eq:hard_bc}.

\begin{figure}[htb]
    \centering
    \tikzset{%
        every neuron/.style={
            circle,
            draw,
            minimum size=0.5cm
        },
        neuron missing/.style={
            draw=none, 
            scale=1,
            execute at begin node=\color{black}$\vdots$
        },
    }
    \begin{tikzpicture}[x=1cm, y=1cm, >=stealth]

        \node at (-0.75, -2.25) [] {$x$};
        \node at (-0.75, -3.25) [] {$y$};
        \node at (4.6, -0.75) [] {$u_x$};
        \node at (4.6, -1.75) [] {$u_y$};
        \node at (4.6, -2.75) [] {$\sigma_{xx}$};
        \node at (4.6, -3.75) [] {$\sigma_{yy}$};
        \node at (4.6, -4.75) [] {$\sigma_{xy}$};

        \foreach \m/\l [count=\y] in {1,2} \node [every neuron/.try, neuron
        \m/.try] (input-\m) at (0,-1.5-\y) {};

        \foreach \m [count=\y] in {1, missing, 2, missing,3} \node [every
        neuron/.try, neuron \m/.try ] (hidden-\m) at (1,-0.9-\y*0.7) {};

        \foreach \m [count=\y] in {1, missing, 2, missing,3} \node [every
        neuron/.try, neuron \m/.try ] (hidden2-\m) at (2,-0.9-\y*0.7) {};

        \foreach \m [count=\y] in {1, missing, 2, missing,3} \node [every
        neuron/.try, neuron \m/.try ] (hidden3-\m) at (3,-0.9-\y*0.7) {};

        \foreach \m [count=\y] in {1,2,3,4,5} \node [every neuron/.try,
        neuron \m/.try ] (output-\m) at (4,0-\y) {};

        \foreach \l [count=\i] in {1,2}
        \draw [<-] (input-\i) -- ++(-1,0);

        \foreach \l [count=\i] in {1,...,5}
        \draw [->] (output-\i) -- ++(1,0);

        \foreach \i in {1,...,2}
        \foreach \j in {1,...,3}
        \draw [->] (input-\i) -- (hidden-\j);

        \foreach \i in {1,...,3}
        \foreach \j in {1,...,3}
        \draw [->] (hidden-\i) -- (hidden2-\j);

        \node at (2.5,-1.6) [] {$\cdots$};
        \node at (2.5,-3) [] {$\cdots$};
        \node at (2.5,-4.4) [] {$\cdots$};

        \foreach \i in {1,...,3}
        \foreach \j in {1,...,5}
        \draw [->] (hidden3-\i) -- (output-\j);
    \end{tikzpicture}
    \caption{Topology of a PINN $\mathcal{N}(\bm{x}; \bm{\theta})$ for linear
    elasticity.}
    \label{fig:pinn_topo}
\end{figure}
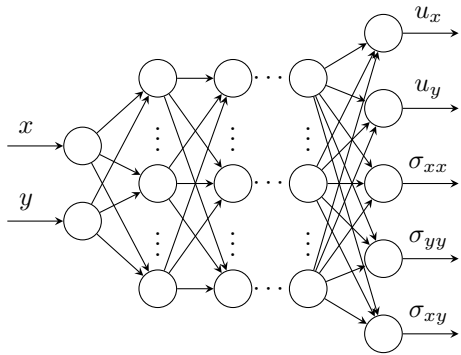

Alternatively, let the ANN output include the displacements $\bm{u}(\bm{x})$ and
the stresses $\bm{\sigma}(\bm{x})$, such that 
\begin{equation}
    \mathcal{N}(\bm{x}; \bm{\theta}) 
    = \left\{ \mathcal{N}_{u_{x}}(\bm{x}; \bm{\theta}), 
        \mathcal{N}_{u_{y}}(\bm{x}; \bm{\theta}),
        \mathcal{N}_{\sigma_{xx}}(\bm{x}; \bm{\theta}), 
        \mathcal{N}_{\sigma_{yy}}(\bm{x}; \bm{\theta}),
    \mathcal{N}_{\sigma_{xy}}(\bm{x}; \bm{\theta})\right\}.
    \label{eq:pinn}
\end{equation}
In the two-dimensional setting this includes two displacement components
$u_x(\bm{x})$, $u_y(\bm{x})$ and three stress components $\sigma_{xx}(\bm{x})$,
$\sigma_{yy}(\bm{x})$, $\sigma_{xy}(\bm{x})$, as the symmetry of the stress
tensor ensures balance of angular momentum \autoref{eq:angular}. This approach
allows the direct computation of the balance law \autoref{eq:lin_mom} without
the usage of second order derivatives. An illustration of the PINN is shown 
in \autoref{fig:pinn_topo}. Furthermore, this topology simplifies the
applicability of hard boundary conditions as explained in \autoref{eq:hard_bc}.
The detailed formulations of the function in \autoref{eq:hard_bc} is case
dependent and will be explained in the experimental section, see
\autoref{sec:Examples}.

\subsection{Loss function and boundary conditions}
\label{sec:loss} \noindent
Using hard boundary conditions, the loss function in \autoref{eq:optimization}
does only contain residual terms. In the two-dimensional setting, these include
\begin{align} 
    \label{eq:loss_el_pinn}
    \mathcal{L}_{PINN}(\bm{x}; \bm{\theta}) &= 
    \mathcal{L}_{\nabla}^x(\bm{x}; \bm{\theta}) 
    + \mathcal{L}_{\nabla}^y(\bm{x}; \bm{\theta}) \\
    &+ \mathcal{L}_{const}^{xx}(\bm{x}; \bm{\theta}) 
    + \mathcal{L}_{const}^{yy}(\bm{x}; \bm{\theta}) 
    + \mathcal{L}_{const}^{xy}(\bm{x}; \bm{\theta}) \nonumber \\
    &+ \mathcal{L}_{W}(\bm{x}; \bm{\theta}), \nonumber 
\end{align}
with
\begin{align*}
    \mathcal{L}_{\nabla}^x(\bm{x}; \bm{\theta}) &= \langle \partial_x
    \mathcal{N}_{\sigma_{xx}} + \partial_y
    \mathcal{N}_{\sigma_{xy}} \rangle, \nonumber \\ 
    \mathcal{L}_{\nabla}^y(\bm{x}; \bm{\theta})
    &= \langle \partial_x \mathcal{N}_{\sigma_{xy}} + \partial_y
    \mathcal{N}_{\sigma_{yy}}\rangle, \nonumber \\
    \mathcal{L}_{const}^{xx}(\bm{x}; \bm{\theta}) &= \langle \lambda
    \left(\partial_x \mathcal{N}_{u_{x}} 
        + \partial_y \mathcal{N}_{u_{y}}
    \right) + 2\mu \partial_x \mathcal{N}_{u_{x}}  
    - \mathcal{N}_{\sigma_{xx}}
    \rangle, \nonumber\\ 
    \mathcal{L}_{const}^{yy}(\bm{x}; \bm{\theta}) &= 
    \langle
    \lambda \left( \partial_x \mathcal{N}_{u_{x}} 
        + \partial_y \mathcal{N}_{u_{y}}
    \right) + 2\mu \partial_y
    \mathcal{N}_{u_{y}} -
    \mathcal{N}_{\sigma_{yy}} \rangle, \nonumber\\
    \mathcal{L}_{const}^{xy}(\bm{x}; \bm{\theta}) &= \langle \mu
    \left( \partial_y \mathcal{N}_{u_{x}}
        + \partial_x \mathcal{N}_{u_{y}}
    \right) - \mathcal{N}_{\sigma_{xy}} \rangle,
    \nonumber \\ 
    \mathcal{L}_{W}(\bm{x}; \bm{\theta}) &= 
    \bigg|\bigg| \displaystyle\sum_{\bm{x} \in \Omega_d}
    \displaystyle\frac{L^2}{2n_d} \mathcal{L}_{W}^{int}(\bm{x}; \bm{\theta}) 
    - \displaystyle\sum_{\bm{x} \in \Gamma_d} \displaystyle\frac{1}{n_b}
    \mathcal{L}_{W}^{ext}(\bm{x}; \bm{\theta}) \bigg|\bigg|_2^2 \nonumber \\
    \mathcal{L}_{W}^{int}(\bm{x}; \bm{\theta}) &= 
    \mathcal{N}_{\sigma_{xx}}\partial_x \mathcal{N}_{u_{x}}
    + \mathcal{N}_{\sigma_{yy}} \partial_y
    \mathcal{N}_{u_{y}}
    +
    \mathcal{N}_{\sigma_{xy}} \left( \partial_y \mathcal{N}_{u_{x}} +
    \partial_x \mathcal{N}_{u_{y}} \right)
    \nonumber \\ 
    \mathcal{L}_{W}^{ext}(\bm{x}; \bm{\theta}) &=
    \mathcal{N}_{\sigma_{xx}}\mathcal{N}_{u_{x}} 
    + \mathcal{N}_{\sigma_{xy}}
    \mathcal{N}_{u_{y}}, \nonumber
\end{align*}
where
\begin{align*}
    \mathcal{N}_{\bullet} = \mathcal{N}_{\bullet}(\bm{x}; \bm{\theta}),\qquad
    \lambda = \lambda(\bm{x}),\qquad \mu = \mu(\bm{x}) \nonumber \\
    \partial_{\star} \bullet= \frac{\partial \bullet}{\partial \star},\qquad
    \langle \bullet \rangle = \displaystyle\sum_{\bm{x} \in \Omega_d}
    \displaystyle\frac{1}{n_d} \bigg|\bigg| \bullet \bigg|\bigg|_2^2.
\end{align*}
Here, $\mathcal{L}_{\nabla}^x(\bm{x}; \bm{\theta})$ and
$\mathcal{L}_{\nabla}^y(\bm{x}; \bm{\theta})$ are the residuals of the balance
law in \autoref{eq:lin_mom}, $\mathcal{L}_{const}^{xx}(\bm{x}; \bm{\theta})$,
$\mathcal{L}_{const}^{yy}(\bm{x}; \bm{\theta})$,
$\mathcal{L}_{const}^{xy}(\bm{x}; \bm{\theta})$ include the constitutive
relation in \autoref{eq:const} and the kinematic relation in \autoref{eq:kin},
whereas $\mathcal{L}_{W}(\bm{x}; \bm{\theta})$ represents the work balance in 
\autoref{eq:energy}. The work balance loss $\mathcal{L}_{W}(\bm{x};
\bm{\theta})$ in \autoref{eq:loss_el_pinn} ensures global conservation, whereas
the other terms represent local laws.

\textit{Remark}: In several works, such as \cite{wang2020understanding} and
\cite{wang2020and}, a weight factor is added to the individual loss terms in
\autoref{eq:loss_el_pinn}. However, in this work no such factors are used, as
numerical experiments carried out by the author did not show any improvement
over equally weighted terms. An exception is the domain decomposition approach
presented in \autoref{sec:cpinn}, where a constant weight factor is used for the
additional interface terms. This
may be related to the usage of full BFGS optimization algorithms instead of
L-BFGS and ADAM and the much simpler loss functions compared to other works,
such as \cite{wang2020understanding} and \cite{wang2020and}.  

\subsection{Loss scaling} \noindent
\label{sec:loss_scaling}
The quantities of the governing equations in \autoref{sec:governing_equations},
namely the spatial location $\bm{x}$, the stress $\bm{\sigma}(\bm{x})$, the
displacement $\bm{u}(\bm{x})$ as well as the material parameters
$\lambda(\bm{x})$ and $\mu(\bm{x})$ \ah{have large differences} in their
respective absolute values. Whereas the displacement is typically in the range
of $u_{\bullet}<< 1$ mm, the material parameters lie in the order of $(\lambda,
\mu)>> 10^3$ MPa. These discrepancies lead to difficulties during optimization
of the loss function $\mathcal{L}_{PINN}$ in \autoref{eq:loss_el_pinn}. To
overcome this problem, the values mentioned above are rescaled internally to lie
in a common order, before calculating the loss terms. This also applies to
boundary conditions. Every quantity is divided by a respective constant, such
that
\begin{equation}
    \hat{\bm{x}} = \frac{\bm{x}}{x_c}, \; 
    \hat{\bm{\sigma}}(\bm{x}) = \frac{\bm{\sigma}(\bm{x})}{\sigma_c}, \;
    \hat{\bm{u}}(\bm{x}) = \frac{\bm{u}(\bm{x})}{u_c}, \;
    \hat{\lambda}(\bm{x}) = \frac{\lambda(\bm{x})}{\lambda_c}, \;
    \hat{\mu}(\bm{x}) = \frac{\mu(\bm{x})}{\mu_c}.
\end{equation}
The rescaled entities used for calculation, denoted by $\hat{\bullet}$, need to
be scaled back again into \ah{its} natural range during prediction, to yield
quantitatively correct results. Typically, the
scaling factors $\bullet_c$ are chosen to be the maximum value of the entity
under consideration with respect to the domain $\Omega$ in \autoref{eq:domain}.
Note that the scaling only \ah{affects} the weighting of the different loss
terms, thus impacts sole the optimization dynamics, but not the dynamics of the
PDE and the spatial derivatives involved. \ah{This is because we consider a
    linear PDE in \autoref{sec:pinn_meca}. The obtained solution fields depend
    qualitatively on the microstructure, the material \ah{parameters'} phase
    contrast of the different materials involved and the specific kind of
    boundary condition in the BVP.  Quantitatively the solution fields depend on
    the values of the single material parameters and the boundary condition
    applied.  Once solved, the solution field can be scaled as needed. This
parametrization enables generalization to different boundary condition values.}

\textit{Remark:} For the sake of simplicity, along this work the unscaled
quantities were used throughout the text and formulae. The rescaling is only
used internally during loss calculation. For large material contrasts, the
proposed PINN method cannot converge without loss scaling.

\subsection{Numerical experiments - PINN}
\label{sec:Examples} \noindent
\ah{In the following,} the proposed method described in \autoref{sec:pinn_meca}
is investigated by several numerical experiments. First, in
\autoref{sec:example_pinn_homo}, a homogeneous case is considered. Second, in
\autoref{sec:example_pinn_inhomo}, the same experimental setup but with a single
inclusion added to the domain is carried out. This turns the BVP into an
inhomogeneous problem, to compare the effect of \rv{heterogeneity} on the
proposed method. If not stated otherwise, all experiments were carried out on a
Nvidia Quadro M5000M GPU using single-precision floating-point format and
TensorFlow 2
\cite{tensorflow2015-whitepaper}.

\subsubsection{Homogeneous plate - PINN}
\label{sec:example_pinn_homo} 
\begin{figure}[htb]
    \centering
    \begin{subfigure}[b]{0.55\textwidth}
        \centering
        \begin{tikzpicture}
            \centering
            \draw[very thick] (0, 0) rectangle (2.75, 2.75);
            \foreach \j in {0,...,4}
            \draw [->, red, very thick] (2.75, 0.6875*\j) -- (3.25, 0.6875*\j);
            \draw [-, red, very thick] (3.25, 2.75) -- (3.25, 0);

            \draw [<-, thick] (0, 1.375) -- (-0.5, 1.375);
            \draw [<-, thick] (1.375, 2.75) -- (1.375, 3.25);
            \draw [<-, thick] (1.375, 0) -- (1.375, -0.5);

            \draw [->, thick] (1.375, 1.375) -- (2.0, 1.375); 
            \draw [->, thick] (1.375, 1.375) -- (1.375, 2.0); 
            \node at (2.0, 1.1) {$x$};
            \node at (1.1, 2.0) {$y$};

            \node at (0.5, 0.5) {$\Omega$};
            \node at (-1.7, 1.35) {${u}_x = {\sigma}_{xy} = 0$};
            \node at (1.375, 3.5) {${\sigma}_{yy} = {\sigma}_{xy} = 0$}; 
            \node at (1.375, -0.75) {${u}_y = {\sigma}_{xy} = 0$};
            \node at (4, 1.35) {$\sigma_{xx} = \bar{{\sigma}}$};
            \node at (4, 1) {$\sigma_{xy} = 0$};
        \end{tikzpicture}
        \label{fig:2d_rect}
        \caption{Homogeneous}
    \end{subfigure}
    \hfill
    \begin{subfigure}[b]{0.35\textwidth}
        \centering
        \begin{tikzpicture}
            \centering
            \draw[very thick] (0, 0) rectangle (2.75, 2.75);
            \foreach \j in {0,...,4}
            \draw [->, red, very thick] (2.75, 0.6875*\j) -- (3.25, 0.6875*\j);
            \draw [-, red, very thick] (3.25, 2.75) -- (3.25, 0);

            \draw[fill] (1.375, 1.375) circle (0.5);

            \node at (0.5, 0.5) {$\Omega$};
            \node at (0, -1.35) {};

    \end{tikzpicture}
        \label{fig:2d_circ}
        \caption{Inhomogeneous}
    \end{subfigure}
    \caption{Rectangular domain $\Omega$ from \autoref{eq:domain} \ah{with
    length $L = 2$ mm} under uniaxial loading \ah{\textit{x}-direction}.
\ah{Plane strain is assumed.} The boundary conditions of (a) also apply to (b).}
    \label{fig:2d}
\end{figure}
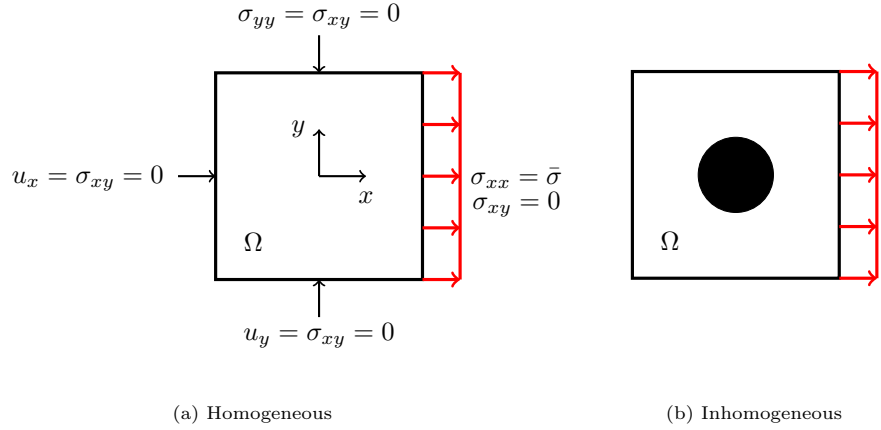

\noindent
The first example considers a two-dimensional \rv{square} domain $\Omega$,
\ah{with dimension $d=2$},
from \autoref{eq:domain} with a homogeneous material distribution shown in
\autoref{fig:2d}(a). The domain is subject to Neumann boundary conditions for
static loading $\bar{\sigma}$ and stress-free boundaries from \autoref{eq:trac}.
Dirichlet displacement boundary conditions from \autoref{eq:disp} ensure that no
free-body movements occur. For this example, a PINN as described in
\autoref{sec:pinn_meca}, with $n_L = 4$ and $n_u = 64$, as defined in
\autoref{eq:ann} and \autoref{eq:layer}, is used. \ah{The
    capacity \ah{increase} of the ANN in terms of number of parameters
    $\bm{\theta}$ can be seen as a form of \textit{p}-refinement. The topology
    presented led
to the best results in preceding experiments, which are not shown in this work.}
The
inputs $\bm{h}^{(0)}$ are
the spatial coordinates $\bm{h}^{(0)} = \bm{x}$ of the domain $\Omega$ from
\autoref{eq:domain} with length $L = 2$ \ah{mm}, such that the domain is a zero
centered \rv{square} between $-1$ and $1$. A traction $\bar{\sigma} = 0.025$ MPa
is applied perpendicular to the boundary, which results in small strains in
accordance to the linear elastic theory. 

\textit{Remark:} In FEM, the zero stress boundary conditions on free
boundaries $\partial \Omega$ are fulfilled automatically due to the weak form
used in the formulation. \ah{In the proposed PINN from \autoref{sec:pinn_meca}},
the strong form of the BVP is used. Therefore it is necessary to explicitly
enforce the zero stress boundary conditions as seen in \autoref{fig:2d}.

The hard boundary conditions from
\autoref{eq:hard_bc} are enforced in consistency with \autoref{fig:2d}(b), such
that
\begin{align}
    {\mathcal{N}}_{u_x}(\bm{x}; \bm{\theta})
    &= 0 + (-1 - x) \tilde{\mathcal{N}}_{u_x}(\bm{x}; \bm{\theta}), \\
    {\mathcal{N}}_{u_y}(\bm{x}; \bm{\theta})
    &= 0 + (-1 - y) \tilde{\mathcal{N}}_{u_y}(\bm{x}; \bm{\theta}),
    \nonumber \\
    {\mathcal{N}}_{\sigma_{xx}}(\bm{x}; \bm{\theta})
    &= \bar{\sigma} + (1 - x) 
    \tilde{\mathcal{N}}_{\sigma_{xx}}(\bm{x}; \bm{\theta}),
    \nonumber \\
    {\mathcal{N}}_{\sigma_{yy}}(\bm{x}; \bm{\theta})
    &= 0 + (1 - y) 
    \tilde{\mathcal{N}}_{\sigma_{yy}}(\bm{x}; \bm{\theta}),
    \nonumber \\
    {\mathcal{N}}_{\sigma_{xy}}(\bm{x}; \bm{\theta})
    &= 0 + \left( (1 - x)^2 (1 - y)^2 \right)
    \tilde{\mathcal{N}}_{\sigma_{xy}}(\bm{x}; \bm{\theta}),
    \nonumber 
\end{align}
The material parameters
$\lambda(\bm{x})$ and $\mu(\bm{x})$ in the constitutive relation
\autoref{eq:const} are calculated
from Young's modulus $E = 1\times10^4$ MPa and Poisson's ratio $\nu = 0.4$. The
collocation points described in \autoref{sec:optimization} are chosen on a
regular grid $\Omega_d$, whereas the number of collocation points is $n_d =
128^2$. For the optimization problem in \autoref{sec:optimization}, the loss
function in \autoref{eq:loss_el_pinn} is minimized by the BFGS algorithm
\cite{fletcher1987practical}, which terminated after 138 iterations due to
gradient size. The prediction is carried out on a regular grid of $n_d = 128^2$
collocation points.

\begin{figure}[H]
    \begin{tikzpicture} 
        \centering 
        \newcommand\X1{0}
        \newcommand\Y1{0}

        \node at (\X1, \Y1 - 0.25) {\includegraphics[width=0.25\textwidth]
        {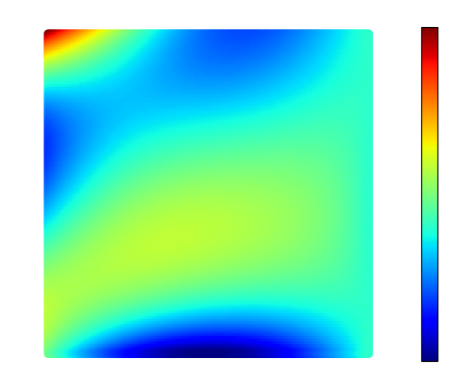}}; \node at (\X1, \Y1 + 1.25)
        {$\sigma_{xx}$ [MPa]}; \node at (\X1 + 2.25, \Y1 + 0.75) {2.50$e-$02};
        \node at (\X1 + 2.25, \Y1 - 1.25) {2.50$e-$02};

        \node at (\X1 + 5, \Y1 - 0.25) {\includegraphics[width=0.25\textwidth]
        {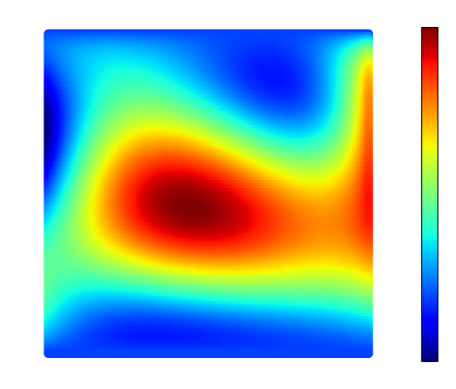}}; \node at (\X1 + 5, \Y1 + 1.25)
        {$\sigma_{yy}$ [MPa]}; \node at (\X1 + 7.25, \Y1 + 0.75) {1.17$e-$06};
        \node at (\X1 + 7.25, \Y1 - 1.25) {-2.72$e-$07};

        \node at (\X1 + 10, \Y1 - 0.25) {\includegraphics[width=0.25\textwidth]
        {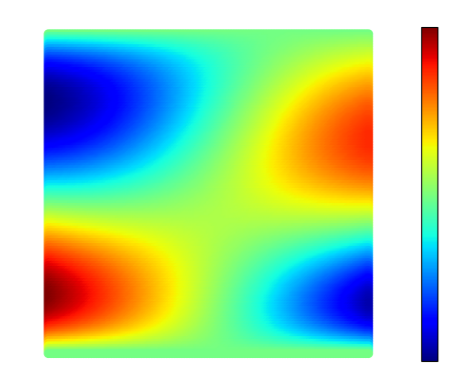}}; \node at (\X1 + 10, \Y1 + 1.25)
        {$\sigma_{xy}$ [MPa]}; \node at (\X1 + 12.25, \Y1 + 0.75) {1.46$e-$06};
        \node at (\X1 + 12.25, \Y1 - 1.25) {-1.40$e-$06};

        \node at (\X1, \Y1 - 3.25) {\includegraphics[width=0.25\textwidth]
        {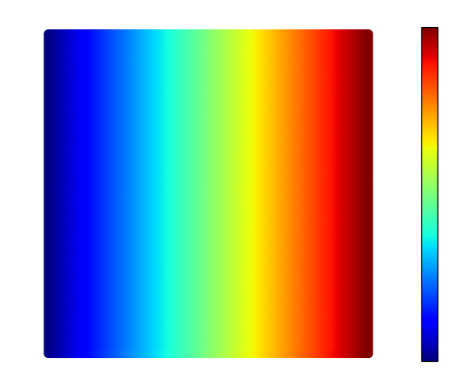}}; \node at (\X1, \Y1 - 1.75)
        {$u_{x}$ [mm]}; \node at (\X1 + 2.25, \Y1 - 2.25) {4.20$e-$06}; \node
        at (\X1 + 2.25, \Y1 - 4.25) {0.00$e+$00};

        \node at (\X1 + 5, \Y1 - 3.25) {\includegraphics[width=0.25\textwidth]
        {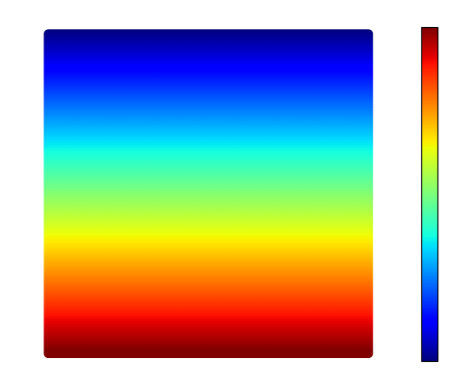}}; \node at (\X1 + 5, \Y1 - 1.75)
        {$u_{y}$ [mm]}; \node at (\X1 + 7.25, \Y1 - 2.25) {0.00$e+$00}; \node
        at (\X1 + 7.25, \Y1 - 4.25) {-2.80$e-$06};

        \node at (\X1 + 10, \Y1 - 3.25)
        {\includegraphics[width=0.25\textwidth]
        {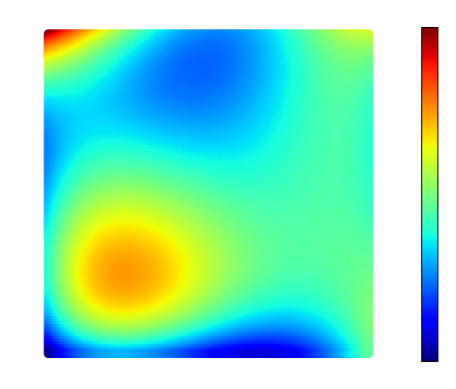}}; 
        \node at (\X1 + 10, \Y1 -1.75) {$W_{int}$ \rv{[--]}}; 
        \node at (\X1 + 12.25, \Y1 - 2.25) {1.05$e-$06}; 
        \node at (\X1 + 12.25, \Y1 - 4.25) {1.05$e-$06}; 

        \node at (\X1, \Y1 - 6.25)
        {\includegraphics[width=0.25\textwidth]
        {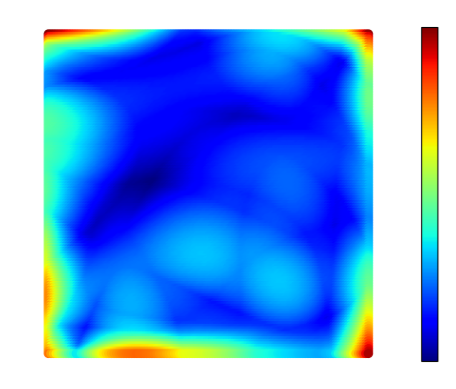}}; 
        \node at (\X1, \Y1 -4.75) {$\mathcal{R}$ \rv{[--]}}; 
        \node at (\X1 + 2.25, \Y1 - 5.25) {4.47$e-$04}; 
        \node at (\X1 + 2.25, \Y1 - 7.25) {1.67$e-$05}; 

        \node at (\X1 + 5, \Y1 - 6.25) 
        {\includegraphics[width=0.25\textwidth]
        {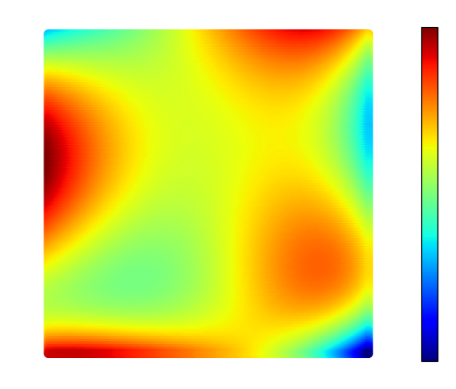}}; 
        \node at (\X1 + 5, \Y1 - 4.75)
        {$\mathcal{R}_{\nabla}^x$ \rv{[--]}}; 
        \node at (\X1 + 7.25, \Y1 - 5.25) {1.04$e-$04}; 
        \node at (\X1 + 7.25, \Y1 - 7.25) {-2.00$e-$04};

        \node at (\X1 + 10, \Y1 - 6.25) 
        {\includegraphics[width=0.25\textwidth]
        {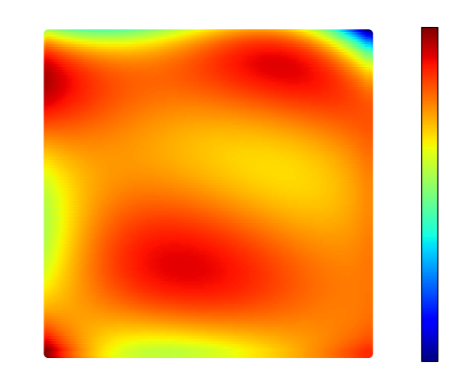}}; 
        \node at (\X1 + 10, \Y1 - 4.75){$\mathcal{R}_{\nabla}^y$
        \rv{[--]}}; 
        \node at (\X1 + 12.25, \Y1 - 5.25) {7.03$e-$05}; 
        \node at (\X1 + 12.25, \Y1 - 7.25) {-2.25$e-$04};

        \node at (\X1, \Y1 - 9.25)
        {\includegraphics[width=0.25\textwidth]
        {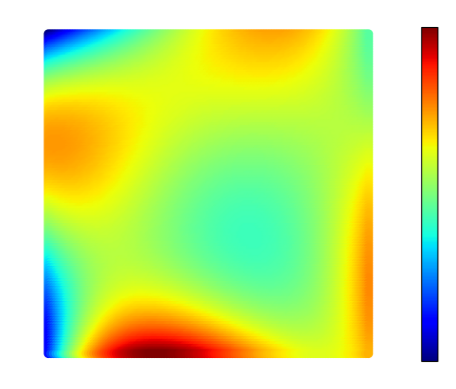}}; 
        \node at (\X1, \Y1 -7.75) {$\mathcal{R}_{const}^{xx}$ \rv{[--]}}; 
        \node at (\X1 + 2.25, \Y1 - 8.25) {1.82$e-$04}; 
        \node at (\X1 + 2.25, \Y1 - 10.25) {-2.50$e-$04}; 

        \node at (\X1 + 5, \Y1 - 9.25)
        {\includegraphics[width=0.25\textwidth]
        {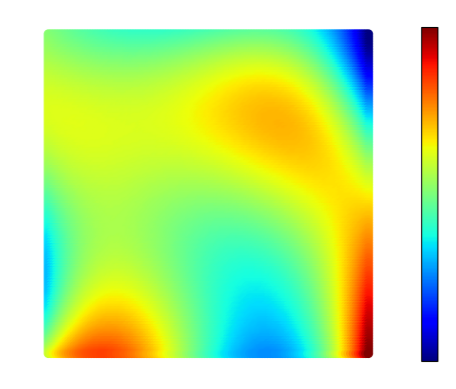}}; 
        \node at (\X1 + 5, \Y1 -7.75) {$\mathcal{R}_{const}^{yy}$ \rv{[--]}}; 
        \node at (\X1 + 7.25, \Y1 - 8.25) {1.05$e-$04}; 
        \node at (\X1 + 7.25, \Y1 - 10.25) {-1.30$e-$04}; 

        \node at (\X1 + 10, \Y1 - 9.25)
        {\includegraphics[width=0.25\textwidth]
        {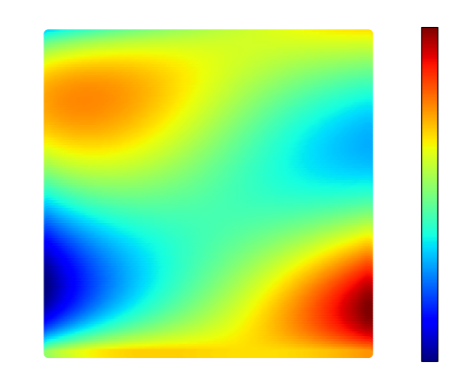}}; 
        \node at (\X1 + 10, \Y1 -7.75) {$\mathcal{R}_{const}^{xy}$ \rv{[--]}}; 
        \node at (\X1 + 12.25, \Y1 - 8.25) {7.23$e-$05}; 
        \node at (\X1 + 12.25, \Y1 - 10.25) {-8.06$e-$05}; 

    \end{tikzpicture} 
    \caption{\textbf{Homogeneous plate - PINN:} The resulting components of the
        stress tensor field $\bm{\sigma}(\bm{x})$ and displacement vector field
        $\bm{u}(\bm{x})$ as well as the point wise internal work $W_{int}$ for
        the homogeneous plate from \autoref{sec:example_pinn_homo} in the top
        rows. In the bottom rows the absolute sum of the residuals $\mathcal{R}$
        as well as the single residuals for the balance law
        $\mathcal{R}_{\nabla}^{\bullet}$ and the constitutive law
        $\mathcal{R}_{const}^{\bullet}$, as described in
        \autoref{sec:example_pinn_inhomo}, are shown. The L2-norm of the work
        balance loss from \autoref{eq:loss_el_pinn} is $|| {\mathcal{L}_W} || =
    2.84221187\times10^{-6}$.}
    \label{fig:exp1_fields} 
\end{figure}

The resulting components of the stress tensor field $\bm{\sigma}(\bm{x})$ and
displacement vector field $\bm{u}(\bm{x})$ as well as the point-wise internal
work $W_{int}(\bm{x})$ are shown in \autoref{fig:exp1_fields}. \ah{Several error
measures are provided,} namely the sum of the absolute
residuals $\mathcal{R}$ as well as the single residuals for the balance law
$\mathcal{R}_{\nabla}^{\bullet} = \{\mathcal{R}_{\nabla}^{x},
\mathcal{R}_{\nabla}^{y}\}$ and the constitutive law
$\mathcal{R}_{const}^{\bullet} = \{\mathcal{R}_{const}^{xx},
\mathcal{R}_{const}^{yy}, \mathcal{R}_{const}^{xy}\}$. They are defined as the
component wise residuals of the balance law in \autoref{eq:lin_mom} for
$\mathcal{R}_{\nabla}^{\bullet}$ and the constitutive relation in
\autoref{eq:const} for $\mathcal{R}_{const}^{\bullet}$, respectively.
$\mathcal{R}$ is defined as the sum of the absolute values
of $\mathcal{R}_{\nabla}^{\bullet}$ and $\mathcal{R}_{const}^{\bullet}$.
\rv{These residuals are obtained for the scaled units, as described in
\autoref{sec:loss_scaling}. Therefore, they are normalized error measures.}

\textit{Remark:} The contour plots for the stresses $\bm{\sigma}(\bm{x})$ and
the point-wise internal work $W_{int}(\bm{x})$ as shown in
\autoref{fig:exp1_fields} show non-homogeneous solution fields. These, however,
deviate only marginally from the correct homogeneous solution, as may be
recognized from the color bar ranges. 

It can be observed, that the PINN achieves low residual errors along the
governing equations. Additionally, the L2-norm of the work balance loss from
\autoref{eq:loss_el_pinn} is $|| \mathcal{L}_W || := \sqrt{\mathcal{L}_W} =
2.84221187\times10^{-6}$, indicating conservation of energy.


\subsubsection{Single inclusion - PINN}
\label{sec:example_pinn_inhomo} \noindent
The second example considers the same two-dimensional \rv{square} domain
$\Omega$ from \autoref{sec:example_pinn_inhomo} with inhomogeneous material
distribution as depicted in \autoref{fig:2d}b. The domain contains a single
circular inclusion and a surrounding matrix phase, whereas the material
parameters for both phases are distinct. For inhomogeneous microstructures
$\Omega$ from \autoref{eq:domain}, the distribution of the material parameters
$\lambda(\bm{x})$ and $\mu(\bm{x})$ in \autoref{eq:const} is usually
non-trivial. To obtain these scalar fields, $\mu$CT-scans are carried out in
practice. These render a voxelized, discrete distribution of different material
densities, from which material phases can be identified. Due to the
discreteness, the transition between different phases are non-smooth. In the
context of PINNs, the optimization problem in \autoref{sec:optimization} becomes
less involved for smoother and more homogeneous problems. To this end, the
distribution of the material parameters $\lambda(\bm{x})$ and $\mu(\bm{x})$ in
\autoref{eq:const} is approximated by smooth functions. In the case of a
\rv{square} domain with a single inclusion, as seen in \autoref{fig:2d}b, the
following equation is a smooth approximation of the material jump
\begin{equation} 
    \lambda(\bm{x} = \{x_1, x_2\}) = c_1 \left(c_2 +
        \operatorname{tanh}\left(\frac{0.4 - \sqrt{x_1^2 +
        x_2^2}}{\delta}\right) \right) + c_3, 
    \end{equation} 
where $c_1$, $c_2$ and $c_3$ are constants manipulating the shape and value of
the inclusion and $\delta$ defines the steepness of the material phase
transition. 

The domain is subject to  Neumann traction boundary conditions for static
loading $\bar{\sigma}$ and stress-free boundaries from \autoref{eq:trac}.
Dirichlet displacement boundary conditions from \autoref{eq:disp} ensure that no
free-body movements occur. For this example, a PINN as described in
\autoref{sec:pinn_meca}, with $n_L = 4$ and $n_u = 64$, as defined in
\autoref{eq:ann} and \autoref{eq:layer}, is used. The inputs $\bm{h}^{(0)}$ are
the spatial coordinates $\bm{h}^{(0)} = \bm{x}$ of the domain $\Omega$ from
\autoref{eq:domain} with length $L = 2$, such that the domain is a zero centered
\rv{square} between $-1$ and $1$. The Neumann boundary condition from
\autoref{eq:trac} is chosen as $\bar{\sigma} = 0.025$ MPa and acts perpendicular
to the boundary. The material parameters $\lambda(\bm{x})$ and $\mu(\bm{x})$ in
the constitutive relation \autoref{eq:const} are calculated from Young's modulus
$E = 1\times10^4$ MPa and Poisson's ratio $\nu = 0.4$ for the inclusion phase
and $E = 1.5 \times 10^3$ MPa and $\nu = 0.4$ for the matrix phase. The
collocation
points are chosen on a regular grid, whereas the number of collocation points is
$n_d = 128^2$. For the optimization problem in \autoref{sec:optimization}, the
loss function in \autoref{eq:loss_el_pinn} is minimized by the BFGS algorithm
\cite{fletcher1987practical}. The optimizer stopped after around $5\times10^3$
iterations due to small gradient updates. The prediction is carried out on a
regular grid of $n_d = 128^2$ collocation points.

\begin{figure}[H] 
    \begin{tikzpicture} 
        \centering 
        \newcommand\X1{0}
        \newcommand\Y1{0}

        \node at (\X1, \Y1 - 0.25) {\includegraphics[width=0.25\textwidth]
        {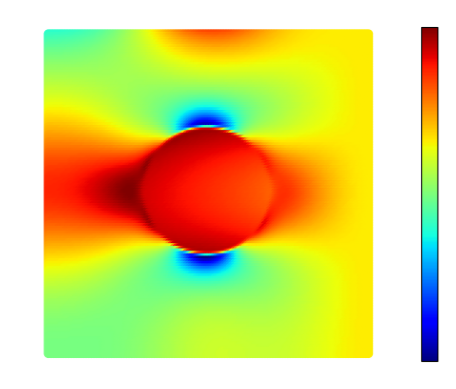}}; 
        \node at (\X1, \Y1 + 1.25){$\sigma_{xx}$ [MPa]}; 
        \node at (\X1 + 2.25, \Y1 + 0.75) {3.52$e-$02};
        \node at (\X1 + 2.25, \Y1 - 1.25) {5.30$e-$03};

        \node at (\X1 + 5, \Y1 - 0.25) {\includegraphics[width=0.25\textwidth]
        {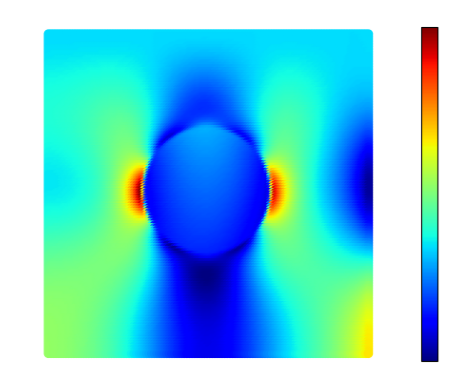}}; 
        \node at (\X1 + 5, \Y1 + 1.25){$\sigma_{yy}$ [MPa]}; 
        \node at (\X1 + 7.25, \Y1 + 0.75) {1.55$e-$02};
        \node at (\X1 + 7.25, \Y1 - 1.25) {-8.00$e-$03};

        \node at (\X1 + 10, \Y1 - 0.25) {\includegraphics[width=0.25\textwidth]
        {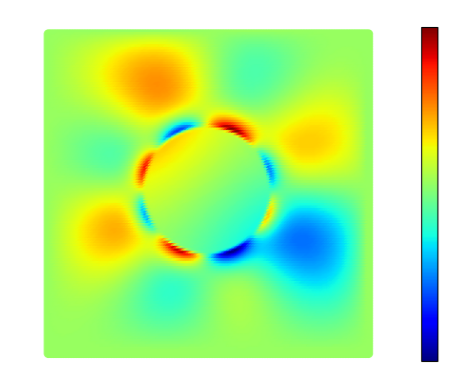}}; 
        \node at (\X1 + 10, \Y1 + 1.25){$\sigma_{xy}$ [MPa]}; 
        \node at (\X1 + 12.25, \Y1 + 0.75) {6.67$e-$03};
        \node at (\X1 + 12.25, \Y1 - 1.25) {-7.63$e-$03};

        \node at (\X1, \Y1 - 3.25) {\includegraphics[width=0.25\textwidth]
        {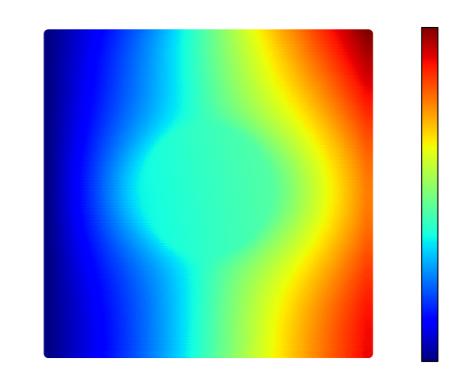}}; 
        \node at (\X1, \Y1 - 1.75){$u_{x}$ [mm]}; 
        \node at (\X1 + 2.25, \Y1 - 2.25) {2.77$e-$05}; \node
        at (\X1 + 2.25, \Y1 - 4.25) {0.00$e+$00};

        \node at (\X1 + 5, \Y1 - 3.25) {\includegraphics[width=0.25\textwidth]
        {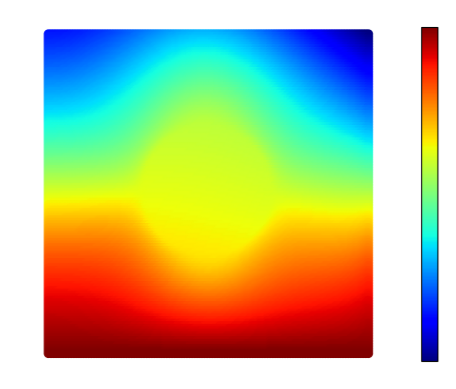}}; 
        \node at (\X1 + 5, \Y1 - 1.75){$u_{y}$ [mm]}; 
        \node at (\X1 + 7.25, \Y1 - 2.25) {0.00$e+$00}; \node
        at (\X1 + 7.25, \Y1 - 4.25) {-1.94$e-$05};

        \node at (\X1 + 10, \Y1 - 3.25)
        {\includegraphics[width=0.25\textwidth]
        {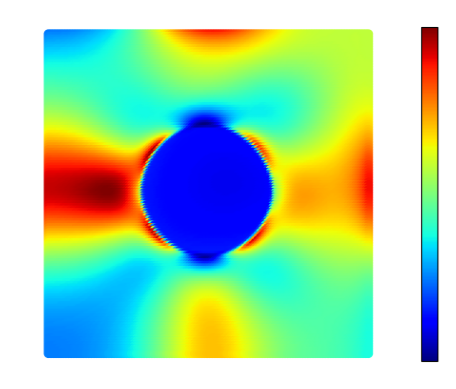}}; 
        \node at (\X1 + 10, \Y1 -1.75) {$W_{int}$ \rv{[--]}}; 
        \node at (\X1 + 12.25, \Y1 - 2.25) {1.15$e-$05}; 
        \node at (\X1 + 12.25, \Y1 - 4.25) {7.22$e-$07}; 

        \node at (\X1, \Y1 - 6.25)
        {\includegraphics[width=0.25\textwidth]
        {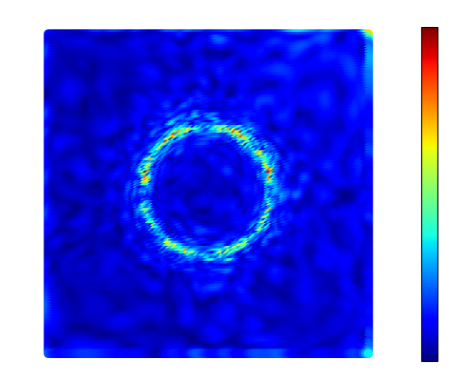}}; 
        \node at (\X1, \Y1 -4.75) {$\mathcal{R}$ [-]}; 
        \node at (\X1 + 2.25, \Y1 - 5.25) {1.78$e-$02}; 
        \node at (\X1 + 2.25, \Y1 - 7.25) {1.60$e-$04}; 

        \node at (\X1 + 5, \Y1 - 6.25) 
        {\includegraphics[width=0.25\textwidth]
        {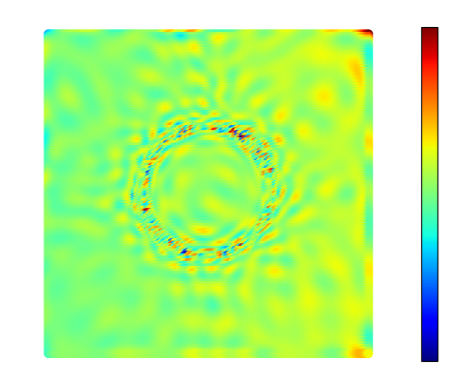}}; 
        \node at (\X1 + 5, \Y1 - 4.75)
        {$\mathcal{R}_{\nabla}^x$ \rv{[--]}}; 
        \node at (\X1 + 7.25, \Y1 - 5.25) {4.32$e-$03}; 
        \node at (\X1 + 7.25, \Y1 - 7.25) {-4.20$e-$03};

        \node at (\X1 + 10, \Y1 - 6.25) 
        {\includegraphics[width=0.25\textwidth]
        {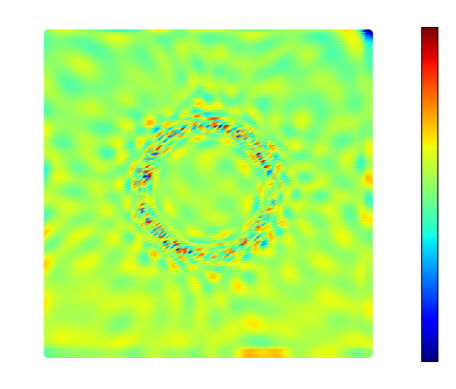}}; 
        \node at (\X1 + 10, \Y1 - 4.75){$\mathcal{R}_{\nabla}^y$
        \rv{[--]}}; 
        \node at (\X1 + 12.25, \Y1 - 5.25) {3.53$e-$03}; 
        \node at (\X1 + 12.25, \Y1 - 7.25) {-4.95$e-$03};

        \node at (\X1, \Y1 - 9.25)
        {\includegraphics[width=0.25\textwidth]
        {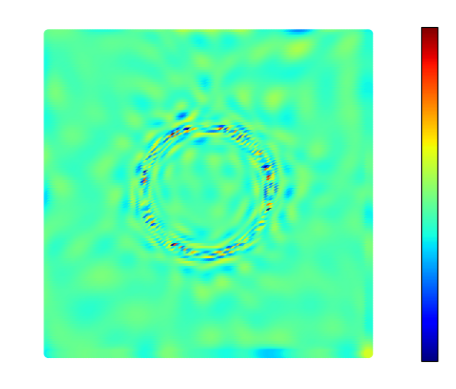}}; 
        \node at (\X1, \Y1 -7.75) {$\mathcal{R}_{const}^{xx}$ \rv{[--]}}; 
        \node at (\X1 + 2.25, \Y1 - 8.25) {6.98$e-$03}; 
        \node at (\X1 + 2.25, \Y1 - 10.25) {-5.67$e-$03}; 

        \node at (\X1 + 5, \Y1 - 9.25)
        {\includegraphics[width=0.25\textwidth]
        {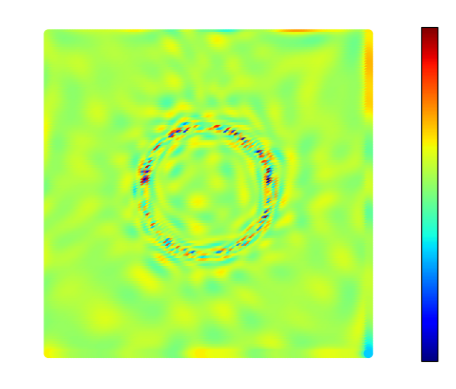}}; 
        \node at (\X1 + 5, \Y1 -7.75) {$\mathcal{R}_{const}^{yy}$ \rv{[--]}}; 
        \node at (\X1 + 7.25, \Y1 - 8.25) {5.02$e-$03}; 
        \node at (\X1 + 7.25, \Y1 - 10.25) {-6.33$e-$03}; 

        \node at (\X1 + 10, \Y1 - 9.25)
        {\includegraphics[width=0.25\textwidth]
        {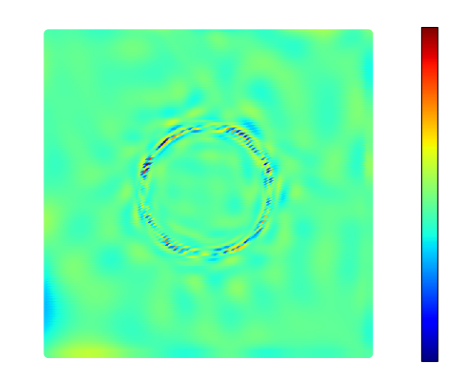}}; 
        \node at (\X1 + 10, \Y1 -7.75) {$\mathcal{R}_{const}^{xy}$ \rv{[--]}}; 
        \node at (\X1 + 12.25, \Y1 - 8.25) {9.22$e-$03}; 
        \node at (\X1 + 12.25, \Y1 - 10.25) {-7.73$e-$03}; 

    \end{tikzpicture} 
    \caption{\textbf{Single inclusion - PINN:} The resulting components of the
        stress tensor field $\bm{\sigma}(\bm{x})$ and displacement vector field
        $\bm{u}(\bm{x})$ as well as the point wise internal work $W_{int}$ for
        the inhomogeneous plate from \autoref{sec:example_pinn_inhomo} in the
        top rows. In the bottom rows the absolute sum of the residuals
        $\mathcal{R}$ as well as the single residuals for the balance law
        $\mathcal{R}_{\nabla}^{\bullet}$ and the constitutive law
        $\mathcal{R}_{const}^{\bullet}$, as described in
        \autoref{sec:example_pinn_homo}, are shown. The L2-norm of the work balance
        loss from \autoref{eq:loss_el_pinn} is $|| \mathcal{L}_W ||  = 4.1985224
    \times 10^{-4}$. Maximum number of BFGS iterations: $1\times10^4$.}
    \label{fig:exp2_fields} 
\end{figure}

The resulting components of the stress tensor field $\bm{\sigma}(\bm{x})$ and
displacement vector field $\bm{u}(\bm{x})$ as well as the point wise internal
work $W_{int}(\bm{x})$ are shown in \autoref{fig:exp2_fields}. The same error
measures as in the previous section are provided, namely the absolute sum of the
residuals
$\mathcal{R}$ as well as the single residuals for the balance law
$\mathcal{R}_{\nabla}^{\bullet} = \{\mathcal{R}_{\nabla}^{x},
\mathcal{R}_{\nabla}^{y}\}$ and the constitutive law
$\mathcal{R}_{const}^{\bullet} = \{\mathcal{R}_{const}^{xx},
\mathcal{R}_{const}^{yy}, \mathcal{R}_{const}^{xy}\}$, \ah{which are defined} in
\autoref{sec:example_pinn_inhomo}. It becomes clear, that the proposed method
captures the inhomogeneity in the material parameters. The L2-norm of the work
balance loss from \autoref{eq:loss_el_pinn} is $|| {\mathcal{L}_W} || =
6.93413196\times10^{-3}$ and the \ah{mean residual is $\mathbb{E}(\mathcal{R}) =
4.1985224 \times10^{-4}$.} The highest errors can be found at the material
interfaces. Here, the maximum residual is $max(\mathcal{R}) = 1.782237
\times10^{-2}$. The minimum residual is $min(\mathcal{R}) = 1.6041845
\times10^{-4}$.  

The PINN is able to qualitatively capture a global solution for the given BVP.
Nevertheless, the phase transition remains problematic due to the sudden jump in
the material parameters $\lambda(\bm{x})$ and $\mu(\bm{x})$ from
\autoref{eq:const}. Therefore, in \autoref{sec:advanched}, two techniques are
studied with the goal to counter this problem.

%

\section{Adaptivity and domain decomposition}
\label{sec:advanched} \noindent
To reduce the error at the phase transitions, two techniques are studied in the
upcoming sections. First, an adaptive algorithm for sampling the collocation
points is proposed in \autoref{sec:adaptive}. Second, a domain decomposition
algorithm to use multiple ANNs on smaller sub-problems is investigated in
\autoref{sec:cpinn}. These are compared to the standard PINN approach in
\autoref{sec:comparison}.

\subsection{Adaptive collocation points}
\label{sec:adaptive} \noindent
To overcome the high errors at the phase transitions reported in
\autoref{sec:example_pinn_inhomo}, an adaptive discretization scheme is
proposed. The adaptive scheme is motivated by FEM \textit{h}-refinement, where a
local mesh refinement at points of interest leads to higher precision in
stress flows. The difference to PINNs is once again, that FEM uses local ansatz
functions, whereas PINNs are global ansatz functions. The goal is, that by
training the PINN on collocation points with high errors, a global solution
representing also \ah{localized} \rv{nonlinearities of the stress and
displacement distributions} is found during optimization of the problem
described in \autoref{sec:optimization}. In the following, the proposed adaptive
algorithm is described. 

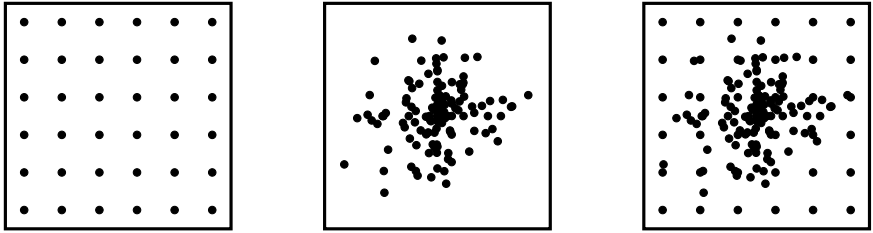
\begin{figure}[htb]
    \centering
    \begin{subfigure}{0.3\textwidth}
        \centering
        \begin{tikzpicture}
            \clip (-1,-1) rectangle (3,3); 
            \draw[very thick] (-0.25, -0.25) rectangle (2.75, 2.75);

            \foreach \x in {0,1,...,5}
            {
                \foreach \y in {0,1,...,5}
                {
                    \node[draw,circle,inner sep=1pt,fill] at (.5*\x,.5*\y) {};
                }
            }
        \end{tikzpicture}
        \caption{regular}
    \end{subfigure}
    \hfill
    \begin{subfigure}{0.3\textwidth}
        \centering
        \begin{tikzpicture}
            \pgfmathsetseed{42}
            \clip (-1,-1) rectangle (3,3); 
            \draw[very thick] (-0.25, -0.25) rectangle (2.75, 2.75);

            \foreach \x in {0,1,...,10}
            {
                \foreach \y in {0,1,...,10}
                {
                    \node[draw,circle,inner sep=1pt,fill] 
                    at (0.125*rand*\x+1.25, 0.125*rand*\y+1.25) {};
                }
            }
        \end{tikzpicture}
        \caption{adaptive}
    \end{subfigure}
    \hfill
    \begin{subfigure}{0.3\textwidth}
        \centering
        \begin{tikzpicture}
            \pgfmathsetseed{42}
            \clip (-1,-1) rectangle (3,3); 
            \draw[very thick] (-0.25, -0.25) rectangle (2.75, 2.75);

            \foreach \x in {0,1,...,5}
            {
                \foreach \y in {0,1,...,5}
                {
                    \node[draw,circle,inner sep=1pt,fill] at (.5*\x,.5*\y) {};
                }
            }

            \foreach \x in {0,1,...,10}
            {
                \foreach \y in {0,1,...,10}
                {
                    \node[draw,circle,inner sep=1pt,fill] 
                    at (0.125*rand*\x+1.25, 0.125*rand*\y+1.25) {};
                }
            }
        \end{tikzpicture}
        \caption{combined}
    \end{subfigure}
    \caption{Two sampling strategies for the choice of collocation points: (a)
    regular grid points as constant basis and (b) adaptive points.} 
    \label{fig:ada_points}
\end{figure}

A related but different approach can be found in \cite{wight2020solving}. First,
a fine regular grid is chosen and the training is carried out as described in
\autoref{sec:example_pinn_inhomo}. This can be seen as a pre-training of the
PINN. Then, for a user defined number of iterations, an adaptive scheme is used
as follows. A set of collocation points $\Omega_d$ is formed by two subsets,
one coarse regular grid $\Omega_d^{reg}$ and one adaptive set $\Omega_d^{rand}$.
The set of regular grid points $\Omega_d^{reg}$ is defined as the Cartesian
product 
\begin{equation}
    \Omega_d^{reg} = \left\{\bm{x}_i \in \left[ -\frac{L}{2}, \frac{L}{2}
        \right] \times \left[ -\frac{L}{2}, \frac{L}{2} \right] ,\; i =
    1,...,n_{reg} \right\}
    \label{eq:fixed_points}
\end{equation}
whereas the random points are chosen randomly from a uniform distribution
\begin{equation}
    \Omega_d^{rand} = \left\{
        \bm{x}_i \sim 
        \bm{\mathcal{U}} \left(-\frac{L}{2}, \frac{L}{2}\right), \; 
    i = 1,...,n_{rand} \right\}
    \label{eq:random_points}
\end{equation}
Then the loss from
\autoref{eq:loss_el_pinn} is calculated for the random points, with the work
balance term neglected, as it is a global, not point-wise loss due to the
appearance of the external work from \autoref{eq:energy}. The loss is
defined as 
\begin{equation}
    \mathcal{L}_{PINN}^{\sim W}(\bm{x}; \bm{\theta}) 
    = \mathcal{L}_{PINN}(\bm{x}; \bm{\theta}) 
    - \mathcal{L}_{W}(\bm{x}; \bm{\theta}).
    \label{eq:loss_ada}
\end{equation}
The set of adaptive points $\Omega_d^{ada}$ are then chosen from the random
points
\begin{equation}
    \Omega_d^{ada} = \left\{\bm{x}_i \in \Omega_d^{rand} \; \bigg| \;
        \mathcal{L}_{PINN}^{\sim W}(\bm{x}_i; \bm{\theta}) \geq
        \mathcal{L}_{PINN}^{\sim W}(\bm{x}_{i+1}; \bm{\theta}), \; i =
        1,...,n_{ada},
    \right\}
    \label{eq:adaptive_points}
\end{equation}
such that a subset of $n_{ada}$ samples with the highest error measures
$\mathcal{L}_{PINN}^{\sim W}(\bm{x}_i; \bm{\theta})$ is formed. The overall set
of collocation points $\Omega_d$ for the adaptive iterations is then formed by
concatenating the regular points $\Omega_d^{reg}$ and the adaptive points
$\Omega_d^{ada}$
\begin{equation}
    \Omega_d = \Omega_d^{reg} \cup \; \Omega_d^{ada}.
    \label{eq:combine}
\end{equation}
Here, the regular grid remains constant during adaptive iterations,
while the adaptive points change at user defined intervals. The \ah{ratio}
$\gamma$ between the number of regular grid points $n_{reg}$ and adaptive grid
points $n_{ada}$ is defined as
\begin{equation}
    \gamma = \frac{n_{reg}}{n_{ada}}.
    \label{eq:gamma}
\end{equation}
The algorithm is summarized in \autoref{algo:adaptivity} and a graphical
representation is shown in \autoref{fig:ada_points}. During training using
the proposed adaptive scheme, the PINN from \autoref{eq:pinn} is trained on
points with large errors. Subsequently, the gradients during the optimization
process in \autoref{eq:optimization} fluctuate widely during the beginning of
each adaptive loop. To avoid exploding gradients and vast alteration of the
already learned weights $\bm{\theta}$ from \autoref{eq:layer}, \textit{gradient
clipping}, as introduced by \cite{szegedy2016rethinking}, is used. The gradients
of the loss function are scaled, if a threshold $\alpha$ is surpassed
\begin{equation}
    \text{if} \; ||\textbf{g}||  > \alpha, \; \text{then} \; \textbf{g}
    \leftarrow \frac{\textbf{g}{\alpha}}{||\textbf{g}||}.
\end{equation}
This stabilizes the training process. Numerical experiments using the adaptive
scheme are carried out in \autoref{sec:comparison}.


\algsetup{indent=2em} 
\begin{algorithm}[htb] 
    \renewcommand{\thealgorithm}{1}
    \caption{Adaptive training of PINN} 
    \label{algo:adaptivity} 
    \begin{algorithmic}[0]
        \FOR{i in $n_{fine}$}
        \STATE Train on fine regular grid $\Omega_d$
        \ENDFOR
        \FOR{i in $n_{iter}$}
        \STATE Create sparse regular grid $\Omega_d^{reg}$ from
        \autoref{eq:fixed_points}
        \STATE Create random points $\Omega_d^{rand}$ from
        \autoref{eq:random_points}
        \STATE Calculate error $\mathcal{L}_{PINN}^{\sim W}(\bm{x}_i;
        \bm{\theta})$ from \autoref{eq:loss_ada} of random points
        $\Omega_d^{rand}$
        \STATE Choose $n_{ada}$ random points with highest error for
        $\Omega_d^{ada}$
        \autoref{eq:adaptive_points}
        \STATE Combine adaptive $\Omega_d^{ada}$ and regular points
        $\Omega_d^{ada}$ \autoref{eq:combine}
        \FOR{i in $n_{ada}$}
        \STATE Train on adaptive and regular points $\Omega_d^{ada}$
        \ENDFOR
        \ENDFOR
    \end{algorithmic} 
\end{algorithm}


\subsection{Conservative PINNs}
\label{sec:cpinn}
\label{sec:cpinn_method} \noindent
The second technique under investigation is called \textit{conservative PINNs}
(\ah{cPINNs}) and utilizes domain decomposition. 

\begin{figure}[htb]
    \centering
    \begin{subfigure}{0.45\textwidth}
        \centering
        \begin{tikzpicture}
            \draw[very thick] (0.0, 0.0) rectangle (4.0, 4.0);
            \node at (0.5, 0.5) {$\Omega$};
        \end{tikzpicture}
        \caption{original}
    \end{subfigure}
    \hfill
    \begin{subfigure}{0.45\textwidth}
        \centering
        \begin{tikzpicture}
            \draw[step=1cm, black, very thick] (0,0) grid (4.0, 4.0);
            \draw[very thick] (0.0, 0.0) rectangle (4.0, 4.0);
            \foreach \x in {0,1,...,3}
            {
                \foreach \y in {0,1,...,3}
                {
                    \node at (0.5+\x, 0.5+\y) {$\Omega_i$};
                }
            }
        \end{tikzpicture}
        \caption{decomposed}
    \end{subfigure}
    \caption{A graphical illustratrion of the domain decomposition from
        \autoref{eq:split}. The original domain $\Omega$ in (a) is decomposed
        \ah{into
    $N$ subdomains $\Omega_i$ in (b).}} 
    \label{fig:domain_split}
\end{figure}
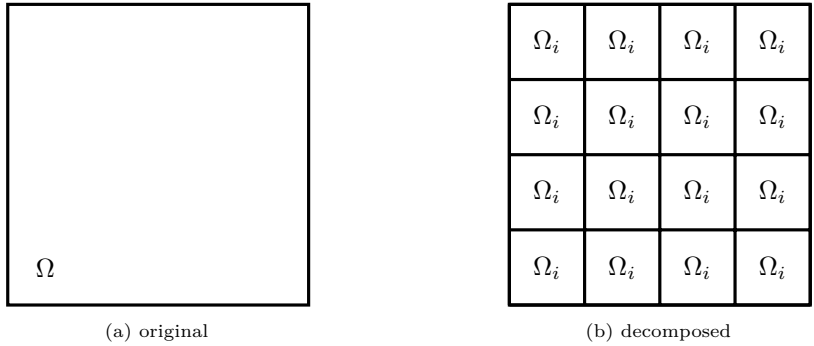

\ah{The use of a global PINN ansatz described in
\autoref{sec:pinn} yields a
\textit{global} solution $\bm{u}(\bm{x})$ in \autoref{eq:ansatz} of the BVP 
in \autoref{eq:bvp}.} This can lead to highly
non-convex optimization landscapes and bad local minima, thus complicating the 
optimization process in \autoref{eq:optimization}. 
To reduce the complexity of the solution, localization by means of a domain
decomposition can be formulated \cite{jagtap2020conservative}. \ah{This can be
seen as a form of \textit{p}-refinement}. In this approach, the domain $\Omega$
from \autoref{eq:bvp} is divided into subdomains 
$\Omega_{i}$, such that
\begin{equation} 
    \Omega = \underset{i=1}{\overset{N}\bigcup} \; \Omega_{i},
    \label{eq:split}
\end{equation} 
where $N$ denotes the number of subdomains $\Omega_{i}$. A graphical
illustration is shown in \autoref{fig:domain_split}. For each subdomain
$\Omega_{i}$, the subdomain solution $\bm{u}_i$ of the BVP is 
provided by a dedicated ANN $\mathcal{N}_i(\bm{x}_i; \bm{\theta}_i)$ with
parameters $\bm{\theta}_i$, such that
\begin{equation}
    \bm{u}_i(\bm{x}_i) \approx \mathcal{N}_i(\bm{x}_i; \bm{\theta}_i), \quad 
    \bm{x}_i \in \Omega_{i}, \quad i = 1, \ldots, N.
    \label{eq:cpinn_ansatz}
\end{equation}
The global solution is obtained by the union of the local solutions
\begin{equation} 
    \bm{u}(\bm{x}) = \underset{i=1}{\overset{N}\bigcup} \; \bm{u}_i(\bm{x}_i)
    \approx \underset{i=1}{\overset{N}\bigcup} \; 
    \mathcal{N}_i(\bm{x}_i; \bm{\theta}_i).
\end{equation} 
To ensure compatibility of the subdomain solutions $\bm{u}_i(\bm{x}_i)$, proper
\textit{interface conditions} on the boundary between adjacent subdomains need
to be introduced. The first condition is solution equality \ah{at} shared
boundary points of two adjacent subdomains
\begin{equation} 
    \bm{u}_i^I(\bm{x}_b) = \bm{u}_i^{II}(\bm{x}_b), \quad 
    x_b \in \Gamma_{I,II} \subset \Omega,
    \label{eq:interface_cond_1}
\end{equation} 
where $\bm{x}_b$ denotes a point on the interface boundary $\Gamma_{I,II}$ of
two subdomains $I$ and $II$. 
The second condition is flux continuity at shared boundary points, which
corresponds to continuous tractions along boundary points in linear elasticity.
Here, these continuity conditions are formulated as
\begin{equation} 
    \bm{\sigma}^{I}(\bm{x}_b) \cdot \bm{n}(\bm{x}_b) 
    = \bm{\sigma}^{II}(\bm{x}_b) \cdot \bm{n}(\bm{x}_b), \quad 
    x_b \in \Gamma_{I,II} \subset \Omega,
    \label{eq:interface_cond_2}
\end{equation} 
\ah{were}, $\bm{\sigma}^{\bullet}(\bm{x}_b)$ denotes the stress along the
interface and $\bm{n}(\bm{x}_b)$ the corresponding normal vector of the boundary
$\Gamma_{I,II}$. The interface conditions from \autoref{eq:interface_cond_1} and
\autoref{eq:interface_cond_2} are then added to the total loss function
$\mathcal{L}$ in \autoref{eq:optimization} in their residual form by means of
soft boundary conditions $\mathcal{L}_{inter}^{u_{\bullet}}$ and
$\mathcal{L}_{inter}^{t_{\bullet}}$,
respectively, such that
\begin{equation}
    \mathcal{L}_{cPINN}(\bm{x}; \bm{\theta}) = \displaystyle\sum_{i=1}^N  
    \mathcal{L}^{net}_{cPINN}(\bm{x}; \bm{\theta})
\end{equation}
with
\begin{align}
    \mathcal{L}^{net}_{cPINN}(\bm{x}; \bm{\theta}) &=  
    \mathcal{L}_{PINN}(\bm{x}; \bm{\theta}) \\
    &+ \psi \mathcal{L}^{u_x}_{inter} (\bm{x}; \bm{\theta})
    + \psi \mathcal{L}^{u_y}_{inter} (\bm{x}; \bm{\theta}) \nonumber \\
    &+ \psi \mathcal{L}_{inter}^{t_x} (\bm{x}; \bm{\theta})
    + \psi \mathcal{L}_{inter}^{t_y} (\bm{x}; \bm{\theta}), \nonumber
    \label{eq:loss_cpinn}
\end{align}
and
\begin{align}
    \mathcal{L}^{u_x}_{inter} (\bm{x}; \bm{\theta}) &= \langle
    \mathcal{N}_{u_x}^I(\bm{x}) - \mathcal{N}_{u_x}^{II}(\bm{x}) \rangle, \:
    \bm{x} \in \Gamma_d^{\textit{hor}} 
    \\
    \mathcal{L}^{u_y}_{inter} (\bm{x}; \bm{\theta}) &= 
    \langle
    \mathcal{N}_{u_y}^I(\bm{x}) - \mathcal{N}_{u_y}^{II}(\bm{x}) \rangle
    , \: \bm{x}
    \in \Gamma_d^{\textit{ver}}
    \nonumber \\
    \mathcal{L}_{inter}^{t_x} (\bm{x}; \bm{\theta}) &= \langle
    \mathcal{N}_{\sigma_{xx}}^I(\bm{x}) 
    - \mathcal{N}_{\sigma_{xx}}^{II}(\bm{x}) \rangle +
    \langle
    \mathcal{N}_{\sigma_{xy}}^I(\bm{x}) 
    - \mathcal{N}_{\sigma_{xy}}^{II}(\bm{x}) \rangle
    , \: \bm{x}
    \in \Gamma_d^{\textit{hor}}
    \nonumber \\
    \mathcal{L}_{inter}^{t_y} (\bm{x}; \bm{\theta}) &= \langle
    \mathcal{N}_{\sigma_{yy}}^I(\bm{x}) 
    - \mathcal{N}_{\sigma_{yy}}^{II}(\bm{x}) \rangle + 
    \langle
    \mathcal{N}_{\sigma_{xy}}^I(\bm{x}) 
    - \mathcal{N}_{\sigma_{xy}}^{II}(\bm{x}) \rangle
    , \: \bm{x}
    \in \Gamma_d^{\textit{ver}}
    \nonumber 
\end{align}
with $\langle \bullet \rangle$ defined over the respective boundaries of the
subnets, similarly to \autoref{eq:loss_el_pinn} and $\bm{x}$ including
$\bm{x}_b$ in a proper sense. Here, $\Gamma_d^{\textit{hor}}$ denotes the
horizontal boundary between two subnets $I$ and $II$, and
$\Gamma_d^{\textit{hor}}$ the vertical one. $\psi$ is a weight factor for the
interface and flux loss to ensure compatibility of the domain.  

The hard boundary conditions from \autoref{eq:hard_bc} have to be adapted to all
ANNs, which share a part of the boundary $\partial \Omega$.


\subsection{Convergence comparison} 
\label{sec:comparison} \noindent
After introducing the adaptive algorithm from \autoref{sec:adaptive} and the
\ah{cPINN} approach from \autoref{sec:cpinn}, in the following numerical
experiment these are compared to the standard PINN approach from
\autoref{sec:pinn_meca} on the single inclusion problem from
\autoref{sec:example_pinn_inhomo}.

To this end, a convergence study with respect to the number of collocation
points is conducted. Four approaches are investigated, where for all approaches
the amount of parameters $\bm{\theta}$ from \autoref{eq:parameters} is kept at
$\bm{\theta} \approx 13 \times 10^3$: 
\begin{enumerate}
    \item
        First, the standard PINN approach, where the number of layers is $n_L =
        4$ and the number of neural units from \autoref{eq:layer} is
        $h^{(l)}_{\eta} = 64$. The PINN is trained for $1 \times 10^4$ BFGS
        iterations.

    \item
        Second, the \ah{cPINN} approach utilizing a 4x4 domain decomposition
        defined in \autoref{eq:split}, using $n_L = 4$ layers $\bm{h}^{(l)}$ and
        an appropriate amount of neural units $h^{(l)}_{\eta}$ to yield the same
        amount of parameters $\bm{\theta}$ as the PINN approach. For every
        subdomain, an appropriate number
of collocation points are chosen on a regular grid, to yield the same number of
total points as the PINN approach. The weight factor $\psi$ from
\autoref{eq:loss_cpinn} is chosen as $\psi = 20$, which showed the best
performance in
pre-studies not shown here. The \ah{cPINN} is trained for $1 \times 10^4$ BFGS
iterations.

    \item
Third, the PINN approach using the proposed adaptive scheme, choosing $n_{fine}
= n_{iter} = 1$ as defined in \autoref{algo:adaptivity}, where each cycle is
trained for $5 \times 10^3$ BFGS iterations. The ratio between adaptive and grid
points $\gamma$, as defined in \autoref{eq:gamma} is chosen as $\gamma = 2.2$,
which yielded the best results. The total number of points is held the same as
the PINN approach. The same topology as the PINN approach described above was
chosen for consistency.  

    \item
        Fourth, the \ah{cPINN} approach using the proposed adaptive scheme,
        choosing $n_{fine} = n_{iter} = 1$ as defined in
        \autoref{algo:adaptivity}, where each cycle is trained for $5 \times
        10^3$ BFGS iterations. The same topology as the \ah{cPINN} approach
        described above was chosen for consistency.  
\end{enumerate}

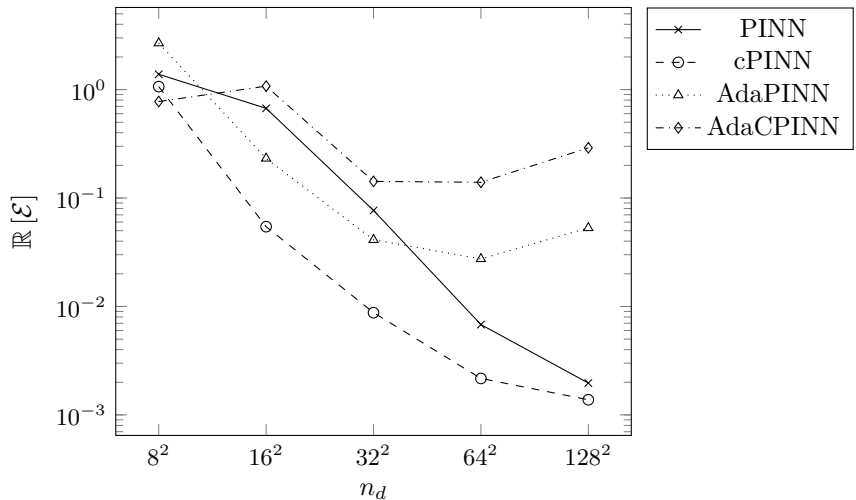
\begin{figure}[H]
    \centering
    \begin{tikzpicture}
        \begin{axis}[ 
                ymode=log,
                xlabel={$n_d$},
                ylabel={$\mathbb{R} \left[\mathcal{E} \right]$},
                legend pos=outer north east,
                xticklabels={{$8^2$}, 
                    {$16^2$}, 
                    {$32^2$}, 
                    {$64^2$}, 
                    {$128^2$}
                },
                xtick={1,...,5},
            ] 
            \addplot[color=black, mark=x] 
            coordinates {
                (1,				1.3845749)
                (2,				0.6716035)
                (3,				0.07682611)
                (4,				0.0068069794)
                (5,				0.0019673028)
            };
            \addlegendentry{PINN}

            \addplot[color=black, dashed, mark=o, mark options=solid] 
            coordinates {
                (1, 1.0655825)
                (2, 0.054440003)
                (3, 0.008768132)
                (4, 0.0021682836)
                (5, 0.0013794703)
            };
            \addlegendentry{\ah{cPINN}}

            \addplot[color=black, dotted, mark=triangle, mark options=solid] 
            coordinates {
                (1, 2.6822066)
                (2, 0.23184381)
                (3, 0.041199565)
                (4, 0.027469713)
                (5, 0.052997183)
            };
            \addlegendentry{AdaPINN}

            \addplot[color=black, dashdotted, mark=diamond, mark options=solid] 
            coordinates {
                (1, 0.77600527)
                (2, 1.0756333)
                (3, 0.14238119)
                (4, 0.13972344)
                (5, 0.2904248)
            };
            \addlegendentry{AdaCPINN}
        \end{axis}
    \end{tikzpicture}
    \caption{\textbf{Convergence comparison:} Calculations of varying number of
        collocation points $n_d$ vs. mean absolute sum of residuals
        $\mathcal{R}$ as defined in \autoref{sec:example_pinn_homo} for the
        different methods under consideration. More precisely, the standard PINN
        approach from \autoref{sec:pinn_meca}, the \ah{cPINN} approach from
        \autoref{sec:cpinn}, and the PINN and \ah{cPINN} approach using adaptive
optimization as described in \autoref{sec:adaptive} \ah{with $\gamma = 2.2$ from
\autoref{eq:gamma}.}}
    \label{fig:exp2_points_graph_mean_residual} 
\end{figure}

Different runs with different maximum number of collocation points where carried
out. More precisely, the maximum number of collocation points where chosen as
$n_d = [8^2, 16^2, 32^2, 64^2, 128^2]$, for all four methods described above.
The error measure is the mean absolute sum of residuals $\mathcal{R}$.
The results are shown in \autoref{fig:exp2_points_graph_mean_residual}.

Both the standard PINN as well as the \ah{cPINN} approach show convergence with
respect to the number of collocation points. Throughout, the \ah{cPINN} approach
yields lower error than the PINN approach. In addition to that, faster
convergence is observed for the \ah{cPINN}. The adaptive variants show
non-convergent behavior from $n_d = 64^2$ total collocation points and upwards.
While the adaptive PINN approach yields slightly lower errors than the standard
PINN approach in the first half of the study, the error raises above the PINN's
in the \ah{second} half. The adaptive \ah{cPINN} approach has the highest error
levels
from all methods for $n_d > 8^2$. \rv{A possible explanation for the
non-convergent behavior of the adaptive approach lies in the total number of
collocation points. From $n_d = 64^2$ total collocation points and upwards, the
number of collocation points is so high, that local refinements does not improve
the solution anymore, as the resolution is fine enough for the underlying
complex solution fields. On the other hand, the performance gets worse, because
the optimization problem becomes more unstable due to the dynamic nature of the
algorithm.}

To summarize, the adaptive scheme is not able to enhance the accuracy of the
solution, at least in the proposed algorithm. It has to be pointed out, that
numerous possibilities to alter the algorithm in \autoref{algo:adaptivity}
exist. The parameters are different ratios $\gamma$ of grid and random points
from \autoref{eq:gamma} for different BVP, different schemes for utilizing grid
optimization and adaptive cycles and different sampling strategies. The domain
decomposition approach on the other hand has less additional parameters and
shows the fastest convergence behavior as well as the lowest mean error. Here,
only one domain decomposition scheme, a 2x2 split, was investigated. The optimal
splitting needs to be defined case by case.

To investigate the performance on a complex microstructure, the \ah{cPINN} from
\autoref{sec:cpinn} with different number of sub-domains is tested on a
$\mu$CT-scan in \autoref{sec:ct_scan}. First, the optimal
splitting of the domain is investigated in \autoref{fig:pre_ct}.

\section{$\mu$CT-scan of a heterogeneous microstructure}
\label{sec:ct_scan} \noindent
This section is concerned with computations on a two-dimensional slice of a
$\mu$CT-scan of a wood-plastic composite (WPC) as \ah{shown} in
\autoref{fig:ct_scan}(a). It consist of two phases, namely a polymer matrix
phase and a wood short fiber inclusion phase. A $\mu$CT-scan yields a voxelized
greyscale image, typically noisy. The noise from the imaging process was
smoothed out by an Gaussian filter, followed by binarization to yield a
voxelized representation of the two-phase material. 

\subsection{Material network}
\label{sec:matnet} \noindent
Unlike in the case of a single inclusion as seen in
\autoref{sec:example_pinn_inhomo}, for real-world microstructures an analytical
description of the distribution of material parameters almost never exists.  For
more complex distributions we therefore propose the usage of an ANN, a
\textit{material network}, which learns the underlying microstructure in a
supervised manner. A densely connected ANN from \autoref{sec:ann} is trained on
the microstructure, such that
\begin{equation} 
    \left( \lambda(\bm{x}), \mu(\bm{x}) \right) \approx
    \mathcal{N}_{mat}(\bm{x}; \bm{\theta}),
\end{equation} 
whereas the activation function $\phi^{(n_L)}$ in
\autoref{eq:hidden} of the last layer $\bm{h}^{(n_L)}$ is defined in the case of
$\lambda(\bm{x})$ as   
\begin{equation} 
    \phi^{(n_L)}(\bm{x}) =
    \left(\operatorname{tanh}(\phi^{(n_L - 1)}(\bm{x})) +
    1\right)\frac{\lambda_{max} - \lambda_{min}}{2} + \lambda_{min},
    \label{eq:matnet} 
\end{equation} 
to only render physically admissible results.
Here, $\lambda_{max}$ and $\lambda_{min}$ are the maximum and minimum values for
$\lambda(\bm{x})$ in the domain $\Omega$ from \autoref{eq:domain}.
\autoref{eq:matnet} is formulated similarly for $\mu(\bm{x})$. \rv{Therefore,
    the material network just learns to predict the correct material phase at a
given spatial coordinate. During training, a pair consisting of a coordinate and
the correct material phase is provided. As no generalization is needed, the
training can be seen as a kind of overfitting to the spatial distribution of
material phases. The resulting material network is a global, infinitely
differentiable or smooth function for the given domain,
which approximates the material distribution. The training usually only lasts
several minutes.}
The resulting
scalar fields then can be used within the PINN. This
approach yields a smooth extension of the microstructure under
consideration. In this work, an ANN  with $n_L = 20$ layers and $n_u = 15$
units trained via BFGS yielded the best results. The material parameters
$\lambda(\bm{x})$ and $\mu(\bm{x})$ are chosen as in
\autoref{sec:example_pinn_inhomo}.

\begin{figure}[htb] 
    \centering
    \begin{subfigure}{0.45\textwidth}
        \includegraphics[height=0.96\textwidth]{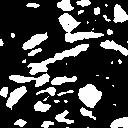}
        \caption{$\mu$CT-scan}
    \end{subfigure}
    \hfill
    \begin{subfigure}{0.45\textwidth}
        \includegraphics[height=1.0\textwidth]
        {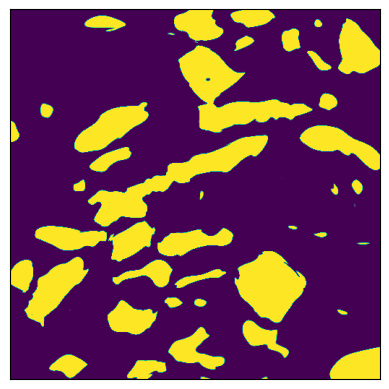}
        \caption{material network}
    \end{subfigure}
    \caption{Microstructure of WPC. (a) shows the $\mu$CT-scan after Gaussian
        filtering and binarization. (b) shows the smooth extension of the
    microstructure by the material network described in \autoref{sec:matnet}.}
    \label{fig:ct_scan}
\end{figure}

A prediction of the material distribution of the $\mu$CT-scan in
\autoref{fig:ct_scan}(a) \ah{is} shown in \autoref{fig:ct_scan}(b). 

\subsection{Numerical experiments} \noindent
The numerical experiment investigating the $\mu$CT-scan in \autoref{fig:ct_scan}
use the same boundary conditions described in
\autoref{sec:example_pinn_homo} and shown in \autoref{fig:2d}. The material
distribution and therefore the distribution of material parameters
$\lambda(\bm{x})$ and $\mu(\bm{x})$ from
\autoref{eq:const} is provided by the material network described in
\autoref{sec:matnet}.  

\subsubsection{Domain split - \ah{cPINN}} 
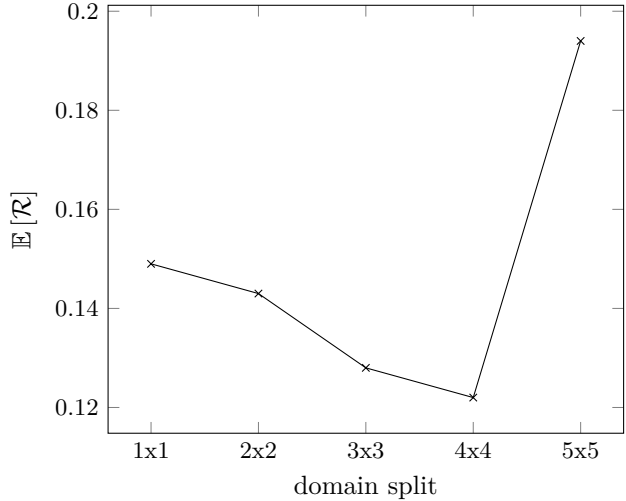
\begin{figure}[H]
    \centering
    \begin{tikzpicture}
        \begin{axis}[ 
                xlabel={domain split},
                ylabel={$\mathbb{E} \left[\mathcal{R} \right]$},
                legend pos=outer north east,
                xticklabels={{1x1}, {2x2}, {3x3}, {4x4}, {5x5}},
                xtick={1,...,5},
            ] 
            \addplot[color=black, mark=x] 
            coordinates {
                (1,				1.49e-1)
                (2,				1.43e-1)
                (3,				1.28e-1)
                (4,				1.22e-1)
                (5,				1.94e-1)
            };
        \end{axis}
    \end{tikzpicture}
    \caption{\textbf{Domain split - \ah{cPINN}:}
        Comparison of different domain splits from \autoref{eq:split} of
        the \ah{cPINN} approach described in \autoref{sec:cpinn}, whereas 1x1
        equals the standard PINN approach from \autoref{sec:pinn_meca}. The
        splits are plotted against the mean of $\mathcal{R}$, the sum of the
        absolute
        residuals described in \autoref{sec:example_pinn_homo}. The 4x4
        \ah{cPINN}
    has a 25\% lower mean residual than the PINN approach.}
    \label{fig:pre_ct}
\end{figure}

\noindent
First, the \ah{cPINN} approach described in \autoref{sec:cpinn} is investigated
with respect to the optimal domain splitting. To this end, several \ah{cPINNs}
with different number $N$ of domain splits $\Omega_i$ from \autoref{eq:split},
but roughly the same amount of parameters $\bm{\theta}$ from
\autoref{eq:parameters}
$\bm{\theta} \approx 13 \times 10^3$, resulting from $n_L = 4$ layers
$\bm{h}^{(l)}$ with the appropriate amount of neural units $h^{(l)}_{\eta}$ from
\autoref{eq:layer}, are trained for $1 \times 10^4$ iterations. The results can
be seen in \autoref{fig:pre_ct}. It can be observed, that the mean of
$\mathcal{R}$, the sum of the absolute residuals described in
\autoref{sec:example_pinn_homo}, is lower for increased number of domain splits
and therefore, with increased number of ANNs, as described in
\autoref{sec:cpinn}. This is explained by the successive localization of
the underlying BVP problem. On the other hand, more loss terms are added to the
optimization problem in \autoref{eq:loss_cpinn}, which renders the optimization
more difficult. This becomes clear in the error jump from the 4x4 split to the
5x5 split in \autoref{fig:pre_ct}. Here, the number of ANNs used raises from 16
to 25 ANNs, a 36\% \ah{increase}, where for every additional ANN the loss term
in \autoref{eq:loss_cpinn} is expanded by the ANNs individual loss terms and
interface conditions. Therefore, the 4x4 split yields the maximum number
of ANNs for the localization advantage to outweigh the complexity gain in
the loss. The 4x4 \ah{cPINN} has a 25\% lower mean residual $\mathcal{R}$
compared to the 1x1 split, which corresponds to the standard PINN approach. 

\subsubsection{$\mu$CT-scan - \ah{cPINN}} 
\begin{figure}[H] 
    \begin{tikzpicture} 
        \centering 
        \newcommand\X1{0}
        \newcommand\Y1{0}

        \node at (\X1, \Y1 - 0.25) {\includegraphics[width=0.25\textwidth]
        {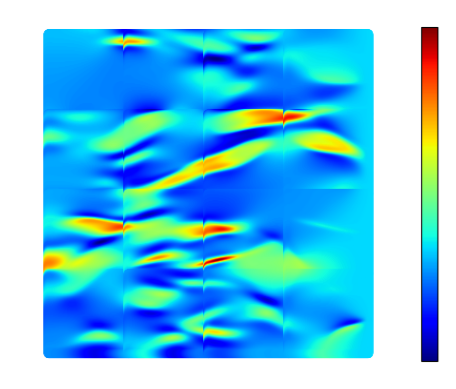}}; 
        \node at (\X1, \Y1 + 1.25){$\sigma_{xx}$ [MPa]}; 
        \node at (\X1 + 2.25, \Y1 + 0.75) {4.94$e-$02};
        \node at (\X1 + 2.25, \Y1 - 1.25) {1.26$e-$02};

        \node at (\X1 + 5, \Y1 - 0.25) {\includegraphics[width=0.25\textwidth]
        {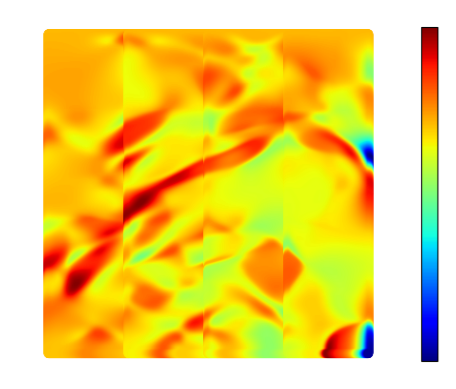}}; 
        \node at (\X1 + 5, \Y1 + 1.25){$\sigma_{yy}$ [MPa]}; 
        \node at (\X1 + 7.25, \Y1 + 0.75) {4.34$e-$03};
        \node at (\X1 + 7.25, \Y1 - 1.25) {-1.09$e-$2};

        \node at (\X1 + 10, \Y1 - 0.25) {\includegraphics[width=0.25\textwidth]
        {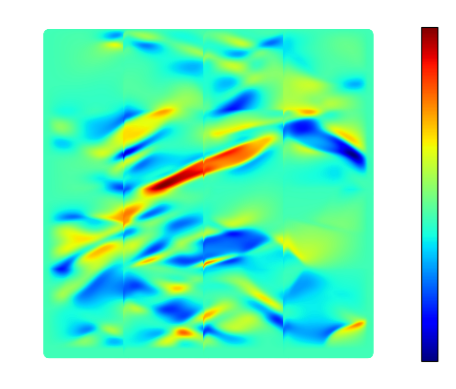}}; 
        \node at (\X1 + 10, \Y1 + 1.25){$\sigma_{xy}$ [MPa]}; 
        \node at (\X1 + 12.25, \Y1 + 0.75) {8.19$e-$03};
        \node at (\X1 + 12.25, \Y1 - 1.25) {-6.12$e-$03};

        \node at (\X1, \Y1 - 3.25) {\includegraphics[width=0.25\textwidth]
        {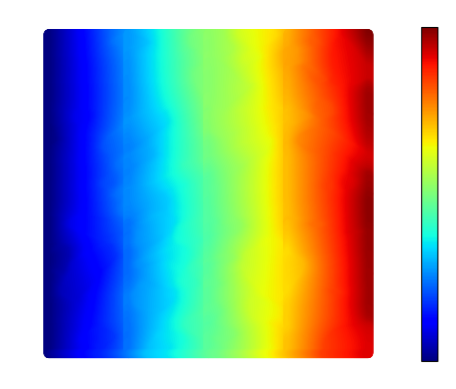}}; 
        \node at (\X1, \Y1 - 1.75){$u_{x}$ [mm]}; 
        \node at (\X1 + 2.25, \Y1 - 2.25) {4.74$e-$06}; \node
        at (\X1 + 2.25, \Y1 - 4.25) {0.00$e+$00};

        \node at (\X1 + 5, \Y1 - 3.25) {\includegraphics[width=0.25\textwidth]
        {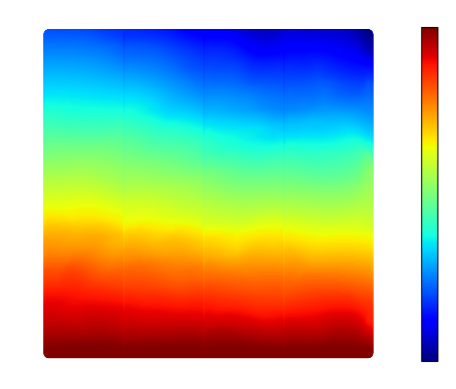}}; 
        \node at (\X1 + 5, \Y1 - 1.75){$u_{y}$ [mm]}; 
        \node at (\X1 + 7.25, \Y1 - 2.25) {0.00$e+$00}; \node
        at (\X1 + 7.25, \Y1 - 4.25) {-1.58$e-$06};

        \node at (\X1 + 10, \Y1 - 3.25)
        {\includegraphics[width=0.25\textwidth]
        {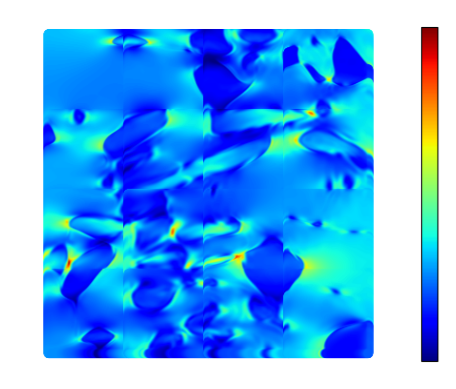}}; 
        \node at (\X1 + 10, \Y1 -1.75) {$W_{int}$ \rv{[--]}}; 
        \node at (\X1 + 12.25, \Y1 - 2.25) {3.66$e-$06}; 
        \node at (\X1 + 12.25, \Y1 - 4.25) {2.44$e-$07}; 

        \node at (\X1, \Y1 - 6.25)
        {\includegraphics[width=0.25\textwidth]
        {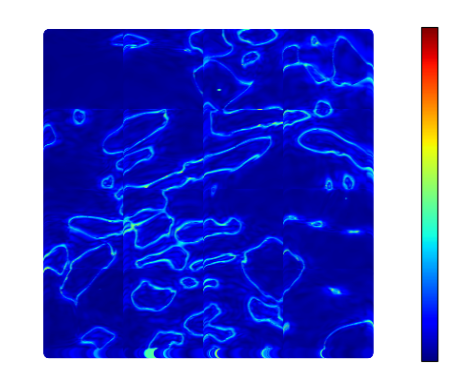}}; 
        \node at (\X1, \Y1 -4.75) {$\mathcal{R}$ [-]}; 
        \node at (\X1 + 2.25, \Y1 - 5.25) {1.71$e+$00}; 
        \node at (\X1 + 2.25, \Y1 - 7.25) {2.97$e-$03}; 

        \node at (\X1 + 5, \Y1 - 6.25) 
        {\includegraphics[width=0.25\textwidth]
        {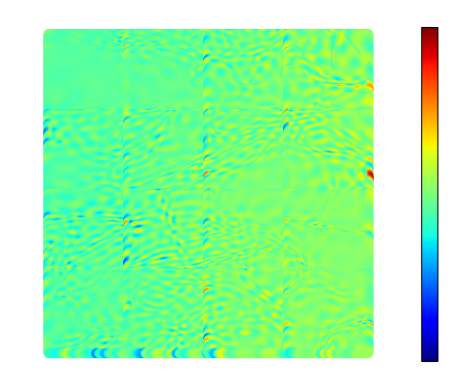}}; 
        \node at (\X1 + 5, \Y1 - 4.75)
        {$\mathcal{R}_{\nabla}^x$ \rv{[--]}}; 
        \node at (\X1 + 7.25, \Y1 - 5.25) {1.94$e-$01}; 
        \node at (\X1 + 7.25, \Y1 - 7.25) {-1.62$e-$01};

        \node at (\X1 + 10, \Y1 - 6.25) 
        {\includegraphics[width=0.25\textwidth]
        {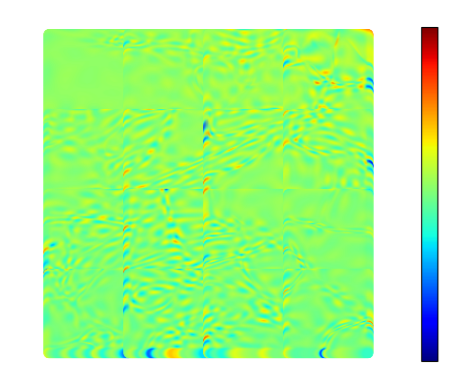}}; 
        \node at (\X1 + 10, \Y1 - 4.75){$\mathcal{R}_{\nabla}^y$
        \rv{[--]}}; 
        \node at (\X1 + 12.25, \Y1 - 5.25) {1.80$e-$01}; 
        \node at (\X1 + 12.25, \Y1 - 7.25) {-1.82$e-$01};

        \node at (\X1, \Y1 - 9.25)
        {\includegraphics[width=0.25\textwidth]
        {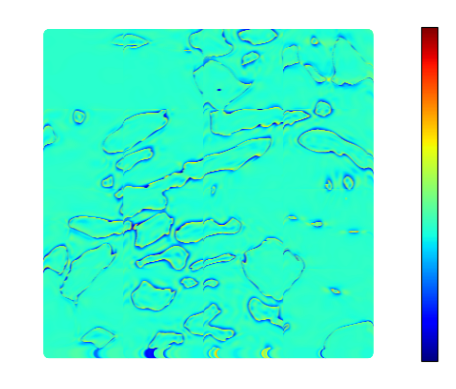}}; 
        \node at (\X1, \Y1 -7.75) {$\mathcal{R}_{const}^{xx}$ \rv{[--]}}; 
        \node at (\X1 + 2.25, \Y1 - 8.25) {1.15$e+$00}; 
        \node at (\X1 + 2.25, \Y1 - 10.25) {-7.87$e-$01}; 

        \node at (\X1 + 5, \Y1 - 9.25)
        {\includegraphics[width=0.25\textwidth]
        {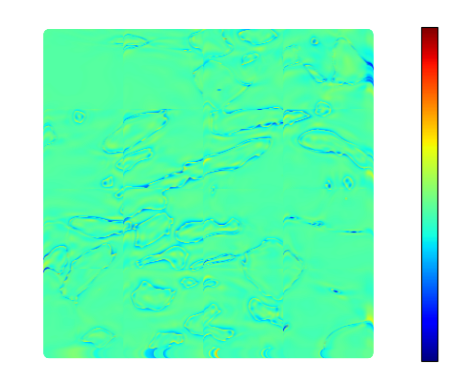}}; 
        \node at (\X1 + 5, \Y1 -7.75) {$\mathcal{R}_{const}^{yy}$ \rv{[--]}}; 
        \node at (\X1 + 7.25, \Y1 - 8.25) {5.81$e-$01}; 
        \node at (\X1 + 7.25, \Y1 - 10.25) {-4.75$e-$01}; 

        \node at (\X1 + 10, \Y1 - 9.25)
        {\includegraphics[width=0.25\textwidth]
        {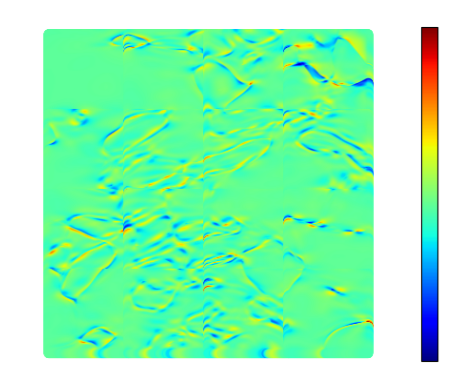}}; 
        \node at (\X1 + 10, \Y1 -7.75) {$\mathcal{R}_{const}^{xy}$ \rv{[--]}}; 
        \node at (\X1 + 12.25, \Y1 - 8.25) {4.58$e-$01}; 
        \node at (\X1 + 12.25, \Y1 - 10.25) {-3.93$e-$01}; 

    \end{tikzpicture} 
    \caption{\textbf{$\bm{\mu}$CT-scan - \ah{cPINN}:}
        The resulting components of the stress tensor field
        $\bm{\sigma}(\bm{x})$ and displacement vector field $\bm{u}(\bm{x})$
        as well as the point wise internal work $W_{int}$ for the
        microstructure from \autoref{sec:ct_scan} using
        cPINNs as described in \autoref{sec:cpinn}, in
        the top rows. In the bottom rows the absolute sum of the residuals
        $\mathcal{R}$ as well as the single residuals for the balance law
        $\mathcal{R}_{\nabla}^{\bullet}$ and the constitutive law
        $\mathcal{R}_{const}^{\bullet}$, as described in
        \autoref{sec:example_pinn_homo}, are shown. The L2-norm of the work
        balance loss from \autoref{eq:loss_el_pinn} is $|| {\mathcal{L}_W} ||
        = 1.6820842 \times10^{-2}$. Maximum number of iterations were $2
        \times 10^4$.
    }
    \label{fig:ct_scan_results} 
\end{figure}
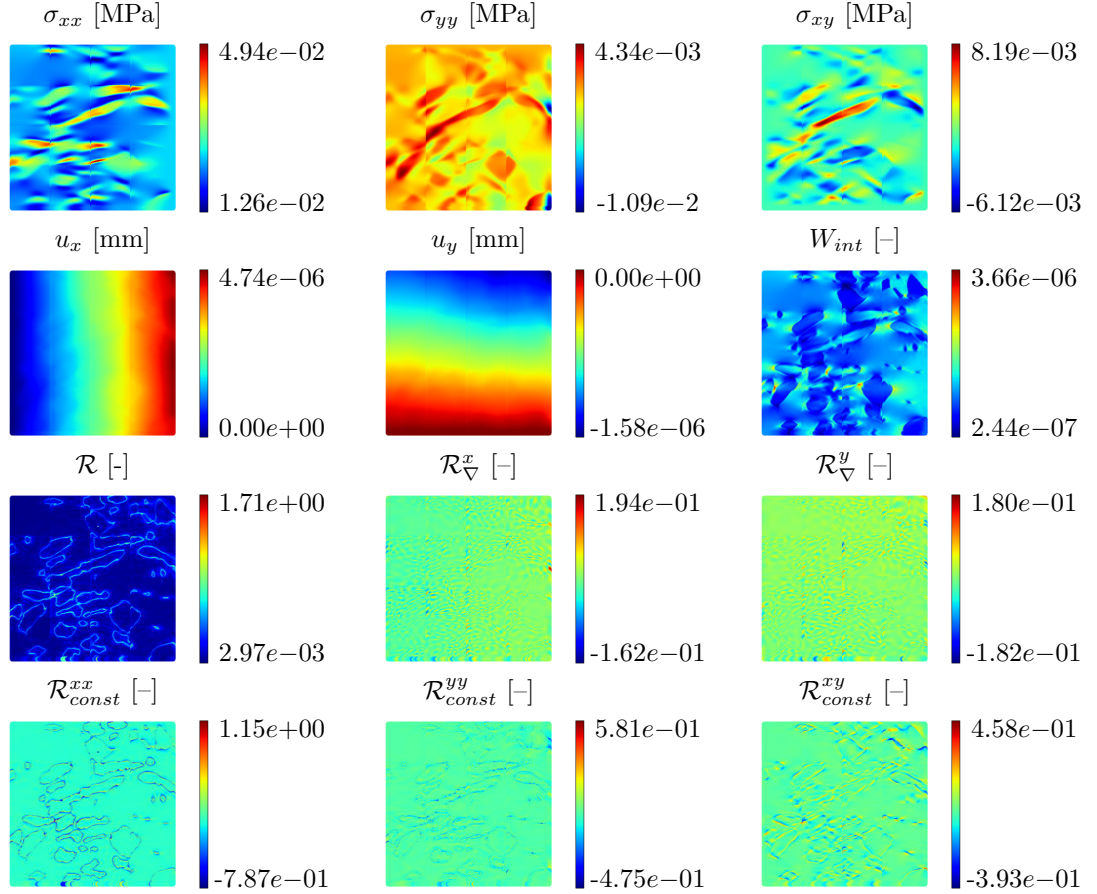

\noindent
Finally, the $\mu$CT-scan is recalculated using the 4x4 \ah{cPINN} with $n_L =
4$ and $n_u = 90$, resulting in $\bm{\theta} \approx 2.7 \times 10^4$ ANN
parameters from \autoref{eq:parameters}. The training as well as the prediction
is carried
out on a regular grid consisting of $n_d = 256^2$ collocation points. A maximum
of $2 \times
10^4$ iterations is used. The results are shown in
\autoref{fig:ct_scan_results}. The maximum of the residual residual
$\mathcal{R}$ from \autoref{sec:example_pinn_homo} is $max(\mathcal{R}) =
9.249407$, the mean $\mathbb{E}(\mathcal{R}) = 1.20354064 \times 10^{-1}$ and
the minimum $min(\mathcal{R}) = 3.2151486 \times 10^{-3}$. The L2-norm of the
work balance loss from \autoref{eq:loss_el_pinn} is $|| {\mathcal{L}_W} || =
1.682084 \times 10^{-2}$. Clearly, the proposed approach is able to resolve the
local tensor, vector and scalar fields of interest, namely the stress tensor
field $\bm{\sigma}(\bm{x})$, the displacement vector field $\bm{u}(\bm{x})$ as
well as the point wise internal work $W_{int}(\bm{x})$. Again, the overall error
$\mathcal{R}$ is highest at the boundaries of the inclusions. Comparing the
residual field $\mathcal{R}$ in \autoref{fig:ct_scan_results} with the
distribution of the material in \autoref{fig:ct_scan}, conformity can be
observed, as the boundaries of the material coincide with the highest error
contours. Even small details are captured. In addition, minor incompatibilities
occur at the subdomain interfaces.


\section{Conclusion and outlook} 
\label{sec:conc_out}\noindent
The objective of this paper is an in-depth study of a physics informed neural
network approach for the solution of the governing equations of linear elastic
continuum micromechanics without the need for training data. PINNs are an
alternative to local numerical methods like FEM. \rv{Additionally, our approach
works directly with the strong form of the elastostatics equations, rendering a
global solution of the underlying BVP.} 

The unique challenges lie in the complex physical
objects under consideration, namely nonlinear tensor fields, which arise due to
the heterogeneous material distribution in the domain. The resulting governing
equations, the balance law, the constitutive equation as well as the kinematic
relation, inherit tensor fields of different scales as well as second order
derivatives, rendering direct optimization difficult. The steep phase transition
between different materials imposes a challenge on gradient based optimization.

In order to overcome these difficulties, a multi-output physics informed neural
network is employed proposed \ah{at \cite{haghighat2021physics}}. The PINN
outputs stress as well as displacement fields, \ah{thereby} circumventing the
need for second order derivatives. An appropriate loss
scaling for the optimization function is proposed, to overcome the problem of
multi-scale loss components, enabling the PINN to
tackle arbitrary material parameter combinations. Boundary conditions of the BVP
are implicitly fulfilled by means of hard boundary conditions, thus simplifying
the loss function and rendering the optimization process more simple.
Furthermore, using hard boundary conditions allows to use smaller ANNs, which
can focus solely on the interior solution fields inside the domain of interest.
The optimization itself is carried out using the BFGS algorithm, which is
computationally feasible due to the small ANN used, yielding faster convergence
rates than ADAM or L-BFGS. To further accelerate the accuracy of the solution,
the work balance of internal and external work is included to the overall loss,
acting like a global constraint besides the point-wise residuals. Furthermore,
an adaptive
sampling method for the collocation points inside the domain is proposed.
Additionally, a domain decomposition cPINN approach proposed in
\cite{jagtap2020conservative} is applied to elastostatics. To
handle arbitrarily complex microstructures, a material network is \ah{employed},
which yields a smooth extension of the underlying material parameter fields,
simplifying the gradient descent optimization by circumventing gradient jumps.

Several numerical examples were given. The first example showed the accuracy of
the standard PINN approach on a homogeneous plate. The BFGS optimizer was able
to find appropriate weights for the PINN to yield state of the art precision. In
a second experiment, the PINN approach was tested qualitatively on an
inhomogeneous plate with a single inclusion. The PINN was able to qualitatively
resolve the underlying tensorial fields. The highest errors appeared in the
phase transition zone. To handle this problem, an adaptive collocation point
sampling algorithm was tested. To further simplify the underlying BVP, a domain
decomposition algorithm was introduced. Comparisons of the standard PINN, the
\ah{cPINN} and adaptive PINN and \ah{cPINN} algorithms showed, that the
\ah{cPINN} approach
using regular sampling yielded the highest accuracy. \rv{The adaptive algorithm
seems to introduce difficulties for the optimization method used. While the
performance was better for low numbers of collocation points, it disoriented for
finer grids.} Finally, the
performance of \ah{cPINN} on a real-world $\mu$CT-scan of WPC was investigated.
Here it was shown, that there is an optimal number of domain splits, which
rendered a trade-off between localization of the problem, thus simplifying the
nonlinear solution fields, and the rising complexity of the loss function, thus
complicating the optimization process. The \ah{cPINN} was able to resolve the
complex stress, displacement and energy fields.

It has been demonstrated that PINNs are able to solve nonlinear partial
differential equations in the context of elastostatics with inhomogeneous
parameters. This problem is challenging for PINNs which approximate the highly
localized solution with a global ansatz. There are several potential
applications for this approach. It was shown that PINNs due to the smoothness of
their solutions and the formulation as an optimization problem are especially
well suited for inverse problems, see e.g., \cite{chen2020physics} and
\cite{chen2021learning}. In this context, the identification of sensor placement
and optimal experimental design could benefit from the adaptive sampling
approach presented in this work. Regarding the identification of material
parameters, physical constraints could render semi-supervised learning possible.
Usually, thousands of data points are needed for conventional supervised
learning approaches. Using PINNs, the regularization by physical laws could
reduce the amount of data needed to yield precise results. \rv{For forward
    problems, the presented PINN approach is not competitive compared to FEM or
    FFT based methods with respect to computational time, as for
    large problems the optimization can last several hours on GPUs. Here, FEM
and FFT methods are more mature. On the other hand, for inverse problems, recent
work in parameter identification of solids are promising
\cite{zhang2020physics}. In inverse problems for continuum mechanics, the loss
is a function of the material parameters, the provided solution fields and the
PDE. The work at hand shows that PINNs are able to resolve the complicated
solutions present in heterogeneous problems. It is therefore \ah{a first step
towards using the PINN approach for regularization of ANNs in inverse problems}
by means of the underlying, known physics, encoded in the PDEs. 
The development of PINN for inverse problems in the context of solid mechanics
is ongoing work of the authors' group \cite{David}.}

Furthermore, towards forward problems, it is possible to include more physics
into the ANN, e.g., more complex material laws or even aleatoric and epistemic
uncertainty in the form of probability distributions or fuzzy methods
\cite{fuhg2021interval}.
Recent developments in parallel \ah{cPINNs} \cite{shukla2021parallel} and
Extended PINNs (XPINNs) \cite{jagtap2020extended} could leverage the
applicability of PINNs to large, multi-scale problems. XPINNs could be an
interesting extension to the \ah{cPINN} approach to increase the accuracy in
microstructural investigations. Here, the phase transition areas are captured by
means of an indicator function. Parallel
PINNs allow the simultaneous calculation of weight updates for multi-network
architectures, making it possible to consider industry-scale problems.

To summarize, the proposed PINN and \ah{cPINN} approaches are able to predict
highly nonlinear tensorial fields in the context of continuum micromechanics.
This work showed, that they are adequately able to capture the underlying
complex physics of microstructural elastostatics. Their full potential is yet to
be investigated, leaving room for further investigations such as uncertainty
quantification and inverse problems.

\clearpage
\section*{Acknowledgement} \noindent
We thank Ameya Jagtap for the fruitful discussion and insights into \ah{cPINNs}
and the Kunststofftechnik Paderborn (KTP) for providing the WPC $\mu$CT-scans.
The support of the research in this work by the German ``Ministerium f\"ur
Kultur
und Wissenschaft des Landes NRW'' is gratefully acknowledged. 

\bibliography{literature_henkes}
\bibliographystyle{elsarticle-num}
\end{document}